\pdfoutput=1

\documentclass[11pt]{article}

\usepackage[final]{acl}

\usepackage{times}
\usepackage{latexsym}

\usepackage[T1]{fontenc}

\usepackage[utf8]{inputenc}

\usepackage{microtype}
\usepackage{color, colortbl}

\usepackage{inconsolata}

\usepackage{graphicx}

\usepackage{multirow}
\usepackage{booktabs}
\usepackage{amssymb}
\usepackage{mathtools}
\usepackage{arydshln}
\usepackage{booktabs}
\usepackage{array}
\usepackage{amsmath}
\usepackage{adjustbox}
\usepackage{xcolor}
\usepackage{dsfont}
\usepackage{xspace}
\usepackage{hyperref}
\newcommand{\llamaduo}{LlamaDuo\xspace} 

\title{\llamaduo: LLMOps Pipeline for Seamless Migration from Service LLMs to Small-Scale Local LLMs}

\author{
Chansung Park\thanks{Equal contributors: Chansung Park, Juyong Jiang, and Fan Wang. Listing order is random.}$^\diamondsuit$,
Juyong Jiang\footnotemark[1]$^\heartsuit$, 
Fan Wang\footnotemark[1]$^\heartsuit$, 
Sayak Paul$^\spadesuit$, 
Jing Tang\thanks{Corresponding author: Jing Tang.}$^{\heartsuit\clubsuit}$ \\ 
$^\diamondsuit$Electronics and Telecommunications Research Institute\\
$^\heartsuit$The Hong Kong University of Science and Technology (Guangzhou) \\
$^\clubsuit$The Hong Kong University of Science and Technology\\
$^\spadesuit$Hugging Face\\
\texttt{\{deep.diver.csp,csjuyongjiang,csfanwang,spsayakpaul\}@gmail.com}\\
\texttt{jingtang@ust.hk}
}

\begin{document}
\maketitle
\begin{abstract}
The widespread adoption of cloud-based proprietary large language models (LLMs) has introduced significant challenges, including operational dependencies, privacy concerns, and the necessity of continuous internet connectivity.  
In this work, we introduce an LLMOps pipeline, ``\llamaduo'', for the seamless migration of knowledge and abilities from service-oriented LLMs to smaller, locally manageable models. This pipeline is crucial for ensuring service continuity in the presence of operational failures, strict privacy policies, or offline requirements.
Our \llamaduo involves fine-tuning a small language model against the service LLM using a synthetic dataset generated by the latter. 
If the performance of the fine-tuned model falls short of expectations, it is automatically improved through additional fine-tuning using extra similar data generated by the service LLM.
This multi-turn process guarantees that the smaller model can eventually match or even surpass the service LLM's capabilities in specific downstream tasks, offering a practical and scalable solution for managing AI deployments in constrained environments.
Extensive experiments with leading-edge LLMs are conducted to demonstrate the effectiveness, adaptability, and affordability of \llamaduo across various downstream tasks. 
Our pipeline implementation is available at \href{https://github.com/deep-diver/llamaduo}{https://github.com/deep-diver/llamaduo}.
\end{abstract}

\section{Introduction}
The emergence of LLMs has significantly transformed a myriad of tasks and domains \cite{chowdhery2023palm,team2023gemini,achiam2023gpt,touvron2023llama,zhao2023survey,jiang2024mixtral,jiang2024survey}.
In particular, cloud-based proprietary LLMs, referred to as service models, such as GPT-4 \cite{achiam2023gpt}, Gemini 1.5 \cite{team2023gemini}, and Claude 3 \cite{claude3}, have exhibited exceptional capabilities when compared to their smaller, open-source counterparts \cite{chang2024survey}. 
A notable survey involving 70 AI industry leaders from diverse enterprises reveals that approximately 80\% of the enterprise market share is dominated by closed-source platforms, with a significant portion of this share attributed to OpenAI \cite{a16zsurvey}.

However, the increasing reliance on cloud-based service models presents significant challenges in terms of operational dependencies \cite{achiam2023gpt}, privacy concerns \cite{wu2024unveiling}, and accessibility challenges \cite{ray2023chatgpt}. 
These challenges manifest in various ways, including potential service disruptions, heightened risks to data privacy due to the transmission of sensitive information to external providers, mandatory internet connectivity for utilization, and inconsistencies stemming from updates to service providers' LLMs \cite{hadi2023survey,zhao2023survey}. 
Additionally, the transition from proof-of-concept (PoC) development utilizing service LLMs to deployment with local models frequently leads to diminished prompt effectiveness owing to differences between models, subsequently resulting in a suboptimal experience for end-users \cite{naveed2023comprehensive,lyu2024keeping}.
To address these concerns and ensure consistent service delivery, it is imperative to develop smaller, locally manageable LLMs that can operate independently of cloud-based infrastructures. 
 
Recent studies have demonstrated that the strategic fine-tuning of smaller and open-source LLMs with high-quality synthetic data \cite{wang2023self,xu2023wizardlm} generated by service LLMs can achieve performances that are on par with, or even surpass, those of proprietary LLMs in specific downstream tasks \cite{vicuna2023,alpaca,luo2023wizardcoder,abdin2024phi,zhou2024lima}. 
Motivated by these findings, we introduce an LLMOps pipeline namely \llamaduo designed to automatically facilitate the seamless migration from service-oriented LLMs to smaller, locally manageable models without the need for human intervention.
Our pipeline begins with utilizing a task-specific initial dataset, referred to as the coverage dataset, to fine-tune a smaller open-source LLM. 
The performance of fine-tuned local LLMs is evaluated using a service LLMs-as-a-Judge strategy \cite{zheng2024judging}. 
If the performance of the fine-tuned model falls short of expectations, we improve it by iteratively fine-tuning on additional synthetic data generated by the service LLM. 
\llamaduo ensures that the smaller model is capable of eventually matching or even surpassing the service LLM’s performance in specific downstream tasks, offering superior long-term economic advantages.
Therefore, it presents a practical and scalable solution for managing AI deployments in environments where resources are limited.
We conduct extensive experiments and analyses across a range of typical tasks, using popular service LLMs such as GPT4o, Claude 3 Sonnet, and Gemini 1.5 Flash, as well as local LLMs, including Gemma 2B and 7B, Mistral 7B, and LLaMA3 8B, to demonstrate that our \llamaduo guarantees the smaller local LLMs possess the potential to eventually match or even exceed the performance of service LLMs in specific downstream tasks.
To summarize, our contributions are as follows:
\begin{itemize}
    \item We introduce \llamaduo, an efficient and affordable LLMOps pipeline designed to facilitate seamless migration from service-oriented LLMs to smaller, locally manageable models without human intervention, ensuring service continuity in constrained environments. 
    \item  
    We employ a multi-turn approach using task-specific synthetic data generated by service LLMs to ensure that \llamaduo empowers the smaller model to eventually match or even exceed the performance of the service LLM in specific downstream tasks.
    \item We substantiate the pipeline's robust performance and adaptability in real-world context through comprehensive experiments across a range of typical tasks, employing popular service LLMs as synthetic data generators and judges for well-known small local LLMs.  
    \item We emphasize the significant economic advantages of \llamaduo for investing in smaller, locally manageable LLMs and their deployment for sustained use, as opposed to the transient benefits derived from the token-based API usage of service LLMs.
\end{itemize}

\section{Related Work}
\subsection{Alignment with Instruction Tuning}
LLMs pretrained on massive corpora demonstrate remarkable capabilities across a wide range of tasks \cite{zhao2023survey,cai2024survey,yoo2024hyperclova,wang2024kasa}. 
Despite their capabilities, a notable challenge with LLMs is their misalignment with user instructions, which limits their practical applications in real-world scenarios \cite{xu2023wizardlm,wang2023self}.
The misalignment stems from the initial pretraining objective of LLMs, which focuses on minimizing generation errors rather than adhering to human instructions \cite{ouyang2022training,chung2024scaling}.
To solve the mismatch, instruction tuning is proposed, which enables LLMs to complete diverse tasks from instructions without significant computational resources or alterations to the model's architecture\cite{longpre2023flan,muennighoff2023crosslingual, taori2023stanford}.
Specifically, instruction tuning involves supplementary training of pretrained LLMs with datasets structured as instruction-output pairs \cite{zhang2023instruction}.
The efficacy of instruction tuning is largely contingent upon the quality and diversity of the instruction datasets employed \cite{wang2024survey}.
However, the process of curating high-quality, diversified data is fraught with challenges, including the extensive time required for creation, privacy concerns, high costs, and the need for substantial human labor \cite{xu2023wizardlm}. 
In response to these challenges, recent studies have explored innovative methods for constructing instruction datasets, notably the utilization of LLMs for data synthesis \cite{liu2024best}.

\subsection{LLM-synthetic Instruction Data}\label{sec:synthetic_data}
LLMs have demonstrated an unprecedented ability to comprehend and execute natural language instructions \cite{ouyang2022training,chung2024scaling,touvron2023llama}. 
This ability is attributed to the process of training LLMs using substantial instruction datasets \cite{wang2023self}. 
However, acquiring massive instruction datasets is challenging due to data scarcity, privacy issues, low data quality, and prohibitive costs associated with manual data curation \cite{abay2019privacy,xu2023wizardlm,liu2024best}. 
Given these constraints, recent studies probe into utilizing LLMs to automatically generate synthetic instruction data \cite{whitehouse2023llmpowered,dai2023auggpt,taori2023stanford}. 
Specifically, these approaches involve prompting powerful LLMs with limited seed data to generate additional synthetic data. 
These data are subsequently employed to fine-tune smaller models, aiming to transfer knowledge to small LLMs and enhance their performance \cite{wang2023let}. 
Leveraging LLMs to generate data can significantly reduce the costs and time for data curation \cite{liu2024best}, while simultaneously improving the efficacy of the fine-tuned models for designated downstream tasks \cite{yang2020generative, puri2020training, guo2023improving, samuel2023can, schlegel2023pulsar}.

\section{LLMOps Pipeline: \llamaduo} 
\begin{figure*}[t]
\centering
\includegraphics[width=0.83\linewidth]{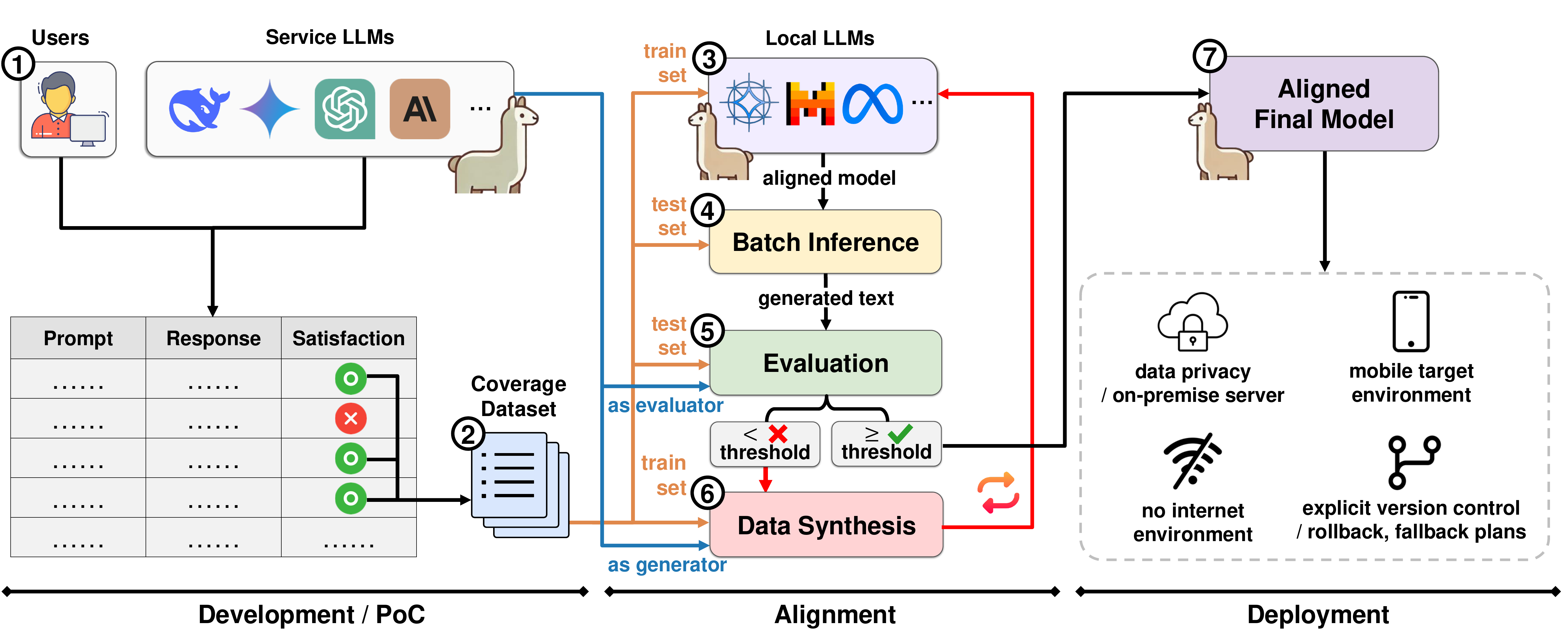}
\vspace{-2mm}
\caption{The LLMOps pipeline namely \llamaduo for migrating from service LLMs to small-scale local LLMs involves three phases. In the Development/PoC phase, \textcircled{1} users manually engineer prompts to interact with service LLMs and \textcircled{2} collect satisfying (prompt, response) pairs into train and test datasets. In the Alignment phase, \textcircled{3} local LLMs are aligned with the train dataset, \textcircled{4} tested on the test dataset, and \textcircled{5} evaluated by service LLMs. \textcircled{6} Synthetic data is generated iteratively until the performance of the aligned model meets a threshold. 
In the Deployment phase, \textcircled{7} the satisfactory model is deployed in constrained environments.}
\label{fig:llamaduo}
\vspace{-5mm}
\end{figure*}

In this section, we elaborate on the details of the proposed \llamaduo, which are depicted in Figure~\ref{fig:llamaduo}.
This LLMOps pipeline aims to ensure service LLMs continuity by transitioning knowledge and abilities from service-oriented LLMs to smaller, locally manageable LLMs without the need for human intervention. 

\subsection{Coverage Dataset}
Users interact with service LLMs through prompt engineering efforts. 
The historical trials composed of the user input prompt and the responses of service LLMs, and potential errors will be recorded and saved in local storage. 
Subsequently, users annotate and collect the most satisfied prompt and response pairs conformed with their real-world use cases. 
The resulting instruction dataset is termed as coverage dataset, denoted as $\mathcal{D}^{(0)} \coloneq \{\mathcal{I}^{(0)}_i, \mathcal{R}^{(0)}_i\}^{N}_{i=1}$, and split as train and test subsets by ratio $\Phi$. Here, $\mathcal{I}^{(0)}_i$ denotes the $i$-th instruction (prompt) in $\mathcal{D}^{(0)}$, $\mathcal{R}^{(0)}_i$ is the corresponding response for the $i$-th instruction, and $N$ is the number of samples in $\mathcal{D}^{(0)}$.
Since coverage dataset is of high quality and satisfying the user's intent in real-world context, the train subsets $\mid\mathcal{D}^{(0)}_{train}\mid=\Phi\cdot N$ will be served as seeds for synthetic datasets generation, while the test subset $\mid\mathcal{D}^{(0)}_{test}\mid=(1-\Phi)\cdot N$ is reserved for performance evaluation of the fine-tuned local LLMs.

\subsection{Fine-tuning}
To efficiently and effectively adapt the local model to specific downstream task(s), we finetune the local LLM with the supervised learning paradigm on high-quality instruction data. 
At the initial cyclicality of the pipeline, the selected local LLM is fine-tuned on the train subsets $\mathcal{D}^{(0)}_{train}$ of the coverage dataset, obtaining the fine-tuned model $\pi^{(0)}$.
At subsequent cyclicality $t$, if the performance of fine-tuned model does not reach or surpass the predetermined evaluation threshold $\mathbf{\varepsilon}$ of specific tasks, the local LLM $\pi^{(t)}$ will be continuously fine-tuned on the increasing number of synthetic data $\{\mathcal{D}^{(1)}_{synth}, \mathcal{D}^{(2)}_{synth}, \ldots, \mathcal{D}^{(t-1)}_{synth}\}$ generated from service LLMs with $\mathcal{D}^{(0)}_{train}$ as seed dataset. Consequently, when $t\ge 1$, the objective of the fine-tuning phase can be formulated as 
\begin{equation}
    \mathcal{L}_{\mathrm{SFT}}(\pi^{(t)}, \mathcal{D}^{(t)}) = -\mathbb{E}
    \left[\log P_{\pi^{(t-1)}}(\mathcal{R}^{(t)}\mid \mathcal{I}^{(t)})\right], 
\end{equation}
where $\mathcal{R}^{(t)}\sim \{\mathcal{D}^{(0)}_{train},\{\mathcal{D}^{(\tau)}_{synth}\}_{\tau=1}^{t-1}\}$ and $\mathcal{I}^{(t)}\sim\mathcal{D}^{(0)}_{train}$.

\subsection{Batch Inference}
After the fine-tuning stage, the fine-tuned local model is prompted with prompts $\mathcal{I}^{(0)}$ sampled from the test subsets $\mathcal{D}^{(0)}_{test}$ of the coverage dataset to produce corresponding response $\mathcal{\hat{R}} \sim \pi^{(t)}(\mathcal{R}^{(0)}\mid \mathcal{I}^{(0)})$. 
To improve the diversity and robustness of responses, the local model generates a batch of $K$ responses $\{\mathcal{\hat{R}}_1, \mathcal{\hat{R}}_2,\ldots,\mathcal{\hat{R}}_K\}$ for each given prompt $\mathcal{I}^{(0)}$. 
Totally, it will construct prompt and responses pairs $\{(\mathcal{I}^{(0)}_i,\mathcal{\hat{R}}_i)\}_{i=1}^{(1-\Phi)\cdot N \cdot K}$. Formally,
\begin{align}
&\mathcal{\hat{R}}_k \sim \pi^{(t)}(\mathcal{R}^{(0)}\mid \mathcal{I}^{(0)}),
\end{align}
where $k\in \{1,2,\ldots, K\}$, $\mathcal{I}^{(0)} \sim \mathcal{D}^{(0)}_{test}$.

\subsection{Evaluation}
In the evaluation stage, we employ ``service LLMs-as-judge'', denoted as $\mathcal{E}_{\mathrm{LLM}}(\cdot)$, to conduct performance evaluation of local model on $\{(\mathcal{I}^{(0)}_i,\mathcal{\hat{R}}_i)\}_{i=1}^{(1-\Phi)\cdot N \cdot K}$. 
Following the works \cite{zheng2024judging,yuan2024self}, the service LLMs evaluate each response triple $(\mathcal{I}^{(0)},\mathcal{\hat{R}},\mathcal{R}^{(0)})$, comprising prompt, the corresponding generated response, and the ground truth, by $M$ times with pairwise comparison and single answer grading strategies.
This evaluation process guarantees the trustworthy and reduces the inherent bias of results. 
Moreover, when employing LLMs as evaluators, the evaluation metrics can be more flexibly adapted to specific tasks, along with a thorough evaluation guide.
In this paper, we measure the similarity between $\mathcal{\hat{R}}$ and $\mathcal{R}^{(0)}$, and how precise $(\mathcal{I}^{(0)},\mathcal{\hat{R}})$ the responses generated by the local LLM answer the given instructions.
These two metrics are provided simultaneously through a prompt, as shown in Figure \ref{fig:default-prompt-assessment} of Appendix \ref{sec:prompt}.
Therefore, $\{(\mathcal{I}^{(0)}_i,\mathcal{\hat{R}}_i,\mathcal{R}^{(0)}_i)\}_{i=1}^{(1-\Phi)\cdot N \cdot K}$ invokes service LLMs to perform evaluation by $(1-\Phi)\cdot N \cdot K \cdot M$ times. 
Subsequently, the evaluation results can be leveraged according to the intention of the operator performing this LLMOps pipeline. 
For example, actions can be taken to increase the reliability of service LLM as an evaluator by calculating the mean or median. 
In this study, we adopt the mean score $V_{\pi^{(t)}}$ and coverage percentage $C_{\pi^{(t)}}$ with $\zeta$ score as evaluation results.  
Here, the coverage percentage $C_{\pi^{(t)}}$ indicates the proportion of responses that have met or exceeded the quality benchmark. 
Formally,\allowdisplaybreaks[4]
\vspace{-2.5mm}
\begin{align}
&V_{\pi^{(t)}} = \frac{1}{(1-\Phi)\cdot N \cdot K}\sum_{j=1}^{(1-\Phi)\cdot N \cdot K}V_{\pi^{(t)}}^j, \\
&C_{\pi^{(t)}} = \frac{1}{(1-\Phi)\cdot N \cdot K}\sum_{j=1}^{(1-\Phi)\cdot N \cdot K}\mathds{1}(V_{\pi^{(t)}}^j\ge \zeta),\\
&V_{\pi^{(t)}}^j = \frac{1}{M}\sum_{m=1}^{M}\mathcal{E}_{\mathrm{LLM}}(\mathrm{prompt}^{(eval)}, d_j), \\
& d_j \sim \{(\mathcal{I}^{(0)}_i,\mathcal{\hat{R}}_i,\mathcal{R}^{(0)}_i)\}_{i=1}^{(1-\Phi)\cdot N \cdot K},
\end{align}
where $V_{\pi^{(t)}}$ and $C_{\pi^{(t)}}$ denote the performance of local LLM at $t$-th cyclicality, $\mathds{1}(\cdot)$ is an indicator function, $\zeta$ denotes a threshold score of $C_{\pi^{(t)}}$, $\mathrm{prompt}^{(eval)}$ is the system prompt used for LLM-as-a-Judge.

\subsection {Data Synthesis}
If the performance of fine-tuned local LLM $V_{\pi^{(t)}}$ or $C_{\pi^{(t)}}$ fails to reach or surpass the predetermined evaluation threshold $\mathbf{\varepsilon}$ of specific tasks, it indicates that fine-tuned local LLM's capabilities are insufficient for the tasks at hand.
Thus, the local LLM cannot yet serve as a substitute for the service LLM and necessitates further refinement.
To achieve this, we utilize service LLMs to generate additional synthetic datasets for fine-tuning local LLM in the next cyclicality. 
To maintain the consistency of data distribution of coverage dataset $\mathcal{D}^{(0)}$ constructed from real-world scenarios, we employ the train subsets $\mathcal{D}^{(0)}_{train}$ as seeds and apply the same framework \cite{wang2023self,alpaca} for synthetic dataset generation.
During synthetic dataset generation, we perform data deduplication to exclude identical samples from $\mathcal{D}^\prime = \{\mathcal{D}^{(0)}_{train}, \{\mathcal{D}^{(1)}_{synth}, \mathcal{D}^{(2)}_{synth}, \ldots, \mathcal{D}^{(t-1)}_{synth}\}\}$ and filter out low-quality samples based on carefully designed rules. 
Finally, we conduct rigorous data decontamination for the synthetic dataset to remove samples that closely resemble those in the test subset $\mathcal{D}^{(0)}_{test}$ of the coverage dataset.
Formally, the data synthesis stage can be formulated as 
\begin{align}
& \mathcal{D}^{(t)}_{synth} \leftarrow \bigcup \psi(\mathcal{D}^{(t)}_{synth},\mathcal{D}^\prime,\mathcal{D}^{(0)}_{test}), \\
&\mathcal{D}^{(t)}_{synth} \sim \mathcal{S}_{\mathrm{LLM}}(\mathrm{prompt}^{(synth)}, seed), \\
& seed \sim \mathcal{D}^{(0)}_{train}, \mathrm{for}\enspace V_{\pi^{(t)}} < \mathbf{\varepsilon}\enspace \mathrm{or}\enspace C_{\pi^{(t)}} < \mathbf{\varepsilon},
\end{align}
where $\bigcup\psi(\cdot,\cdot,\cdot)$ represent a series of data post-processing operations, $\mathcal{D}^{(t)}_{synth}$ denotes synthetic data generated from service LLMs at $t$-th cyclicality, $\mathcal{S}_{\mathrm{LLM}}$ and $\mathrm{prompt}^{(synth)}$ are the service LLM and system prompt used for the data synthesis, respectively.

\section{Experiments}
In this section, we present a comprehensive evaluation of our \llamaduo across a series of settings, demonstrating its robust performance and adaptability in real-world scenarios. 

\subsection{Experimental Settings}
\textbf{Tasks and coverage dataset.}
We select four categories of downstream tasks-summarization, classification, coding, and closed QA-based on their prevalent use and relevance to the operational scope of service LLMs.
We utilize the open-source ``No Robots'' \cite{no_robots} dataset as the coverage dataset. 
This coverage dataset consists of 10K high-quality prompt and response pairs across 10 categories, crafted by expert annotators.
Specifically, we utilize four subsets of the coverage dataset, each corresponding to our targeted tasks.
These subsets serve as seeds for generating synthetic data that can closely align with user expectations for LLM interactions.

\noindent\textbf{Service and local LLMs.}
Considering the API cost effectiveness, rate limit, and model utility, we select popular service LLMs including GPT4o by OpenAI, Claude 3 Sonnet by Anthropic, and Gemini 1.5 Flash by Google to serve as synthetic data generators and judges. 
As for the small-scale local LLMs to be fine-tuned, we opt for the open-source Gemma 2B and 7B \cite{team2024gemma}, Mistral 7B \cite{jiang2023mistral}, and LLaMA3 8B \cite{meta2024introducing} as the base models. 
This selection is motivated by our aim to rigorously evaluate the efficacy and adaptability of our proposed pipeline across diverse settings. The varying scales of base models facilitate a nuanced comparison, allowing us to assess the impact of model scale on performance improvements.
However, as a model-agnostic LLMOps pipeline, our \llamaduo can be generalized to various forms of service and local LLMs beyond the aforementioned models.

\begin{table*}[t]
\centering
\caption{Performance of the service LLMs and local LLMs fine-tuned on $128$K synthetic dataset produced by GPT4o, evaluated by GPT4o, Claude 3 Sonnet, and Gemini 1.5 Flash as judges on test subsets of coverage dataset. 
Each entry is presented as mean score / coverage percentage (\%) with 50 score / coverage percentage (\%) with 70 score. The best results from service and local LLMs are highlighted in \textbf{bold}. 
``\textbf{P-Match}'' represents performance matching, which is defined as the best performance of the local LLM divided by the best performance of the service LLM, with the best results highlighted in \textbf{bold} across different judges.
}
\vspace{-0.5mm}
\resizebox{\linewidth}{!}{
\begin{tabular}{cccccccc}
\toprule
\multirow{2}{*}{{\textbf{Task}}} & \multirow{2}{*}{{\textbf{Model}}} 
& \multicolumn{2}{c} {{\textbf{GPT4o}}} & \multicolumn{2}{c}{{\textbf{Claude 3 Sonnet}}} & \multicolumn{2}{c}{{\textbf{Gemini 1.5 Flash}}}\\
\cmidrule(lr){3-4} \cmidrule(lr){5-6} \cmidrule(lr){7-8}
& ~
& \textbf{Precision$\uparrow$} & \textbf{Similarity$\uparrow$} 
& \textbf{Precision$\uparrow$} & \textbf{Similarity$\uparrow$} 
& \textbf{Precision$\uparrow$} & \textbf{Similarity$\uparrow$} 
\\
\midrule
\multirow{8}{*}{Summarization}  
& GPT4o 
& \textcolor{black}{\textbf{90.71}} / \textcolor{black}{\textbf{97 \%}} / \textcolor{black}{\textbf{96\%}} 
& \textcolor{black}{\textbf{82.00}} / \textcolor{black}{\textbf{95\%}} / \textcolor{black}{\textbf{89\%}} 
&  93.25 / \textcolor{black}{\textbf{100\%}} / \textcolor{black}{\textbf{100\%}} 
& \textcolor{black}{\textbf{86.60}} / \textcolor{black}{\textbf{100\%}} / \textcolor{black}{\textbf{95\%}}
& \textcolor{black}{\textbf{87.10}} / \textcolor{black}{\textbf{100\%}} /  92\%
& \textcolor{black}{\textbf{67.45}} / 85\% / \textcolor{black}{\textbf{48\%}} \\

& Claude 3 Sonnet 
& 88.04 / \textcolor{black}{\textbf{97\%}} / 92\%
& 78.18 / \textcolor{black}{\textbf{95\%}} / 78\%
& \textcolor{black}{\textbf{93.39}} / \textcolor{black}{\textbf{100\%}} / 99\%
& 85.55 / \textcolor{black}{\textbf{100\%}} / \textcolor{black}{\textbf{95\%}}
& 86.70 / \textcolor{black}{\textbf{100\%}} / 92\% 
& 64.10 / 80\% / 36\% \\

& Gemini 1.5 Flash 
& 87.90 / 96\% / \textcolor{black}{\textbf{96\%}}
& 79.14 / \textcolor{black}{\textbf{95\%}} / 88\%
& 91.95 / \textcolor{black}{\textbf{100\%}} / 98\% 
& 85.05 / \textcolor{black}{\textbf{100\%}} / \textcolor{black}{\textbf{95\%}}
& 85.65 / 98\% / \textcolor{black}{\textbf{96\%}}
& 66.45 / \textcolor{black}{\textbf{89\%}} /  40\% \\
\cdashline{2-8}

& Gemma 2B 
& 57.60 / 64\% / 35\%
& 54.49 / 61\% / 35\%
& 74.89 / 86\% / 69\%
& 64.09 / 73\% / 50\%
& 61.90 / 78\% / 40\%
& 42.15 / 38\% / 12\% \\

& Gemma 7B 
& 73.54 / 85\% / 65\%
& 68.58 / 85\% / 59\%
& 86.19 / \textcolor{black}{\textbf{99\%}}  / 93\%
& 77.41 / 94\%  / 77\%
& \textcolor{black}{\textbf{74.59}} / \textcolor{black}{\textbf{95\%}} / \textcolor{black}{\textbf{69\%}}
& \textcolor{black}{\textbf{53.92}} / \textcolor{black}{\textbf{65\%}} / 22\% \\

& Mistral 7B 
& \textcolor{black}{\textbf{76.38}} / \textcolor{black}{\textbf{93\%}} / 70\% 
& 69.65 / \textcolor{black}{\textbf{88\%}}  / 56\%
& 86.20 / \textcolor{black}{\textbf{99\%}} / 92\%
& \textcolor{black}{\textbf{78.44}} / \textcolor{black}{\textbf{96\%}} / 80\% 
& 72.74 / \textcolor{black}{\textbf{95\%}} / 62\% 
& 50.15 / 54\% / 14\% \\

& LLaMA3 8B 
& 75.67 / 88\% / \textcolor{black}{\textbf{75\%}} 
& \textcolor{black}{\textbf{70.54}} / 86\% / \textcolor{black}{\textbf{69\%}} 
& \textcolor{black}{\textbf{87.02}} / \textcolor{black}{\textbf{99\%}} / \textcolor{black}{\textbf{94\%}} 
& 78.42 / 93\% / \textcolor{black}{\textbf{86\%}} 
& 72.74 / 91\% / 64\% 
& 52.23 / 64\% / \textcolor{black}{\textbf{25\%}} \\

& \cellcolor[gray]{0.93}\textbf{P-Match}$\uparrow$ 
& \cellcolor[gray]{0.93}84.20\% / {95.88\%} / 78.13\%
& \cellcolor[gray]{0.93}86.02\% / {92.63\%} / 77.53\% 
& \cellcolor[gray]{0.93}\textcolor{blue}{\textbf{93.18\% / 99\%}} / \textcolor{blue}{\textbf{94\%}} 
& \cellcolor[gray]{0.93}\textcolor{blue}{\textbf{90.58\% / 96\%}} / \textcolor{blue}{\textbf{90.53\%}}
& \cellcolor[gray]{0.93}85.64\% / 95\% / 71.88\% 
& \cellcolor[gray]{0.93}79.94\% / 73.03\% / 52.08\% \\

\midrule
\multirow{8}{*}{Classification}  
& GPT4o 
& 83.62 / \textcolor{black}{\textbf{94\%}} / 81\%
& \textcolor{black}{\textbf{74.45}} / 80\% / 66\%
& 87.50 / 92\% / 92\%
& 72.28 / 72\% / 66\%
& 82.68 / 94\% / 80\% 
& 63.06 / 67\% / 44\% \\

& Claude 3 Sonnet 
& 82.32 / 92\% / 78\%
& 71.53 / \textcolor{black}{\textbf{81\%}} / 70\%
& \textcolor{black}{\textbf{92.89}} / \textcolor{black}{\textbf{100\%}}  / \textbf{100\%}
& 75.07 / \textcolor{black}{\textbf{81\%}} / 73\%
& \textcolor{black}{\textbf{87.34}} / \textcolor{black}{\textbf{97\%}} / \textbf{97\%}
& \textcolor{black}{\textbf{67.18}} / \textcolor{black}{\textbf{80\%}} / 45\% \\

& Gemini 1.5 Flash 
& \textcolor{black}{\textbf{85.43}} / \textcolor{black}{\textbf{94\%}} / \textbf{91\%}
& 72.73 / \textcolor{black}{\textbf{81\%}} / \textbf{75\%}
& 89.03 / 94\% / 89\%
& \textcolor{black}{\textbf{77.96}} / \textcolor{black}{\textbf{81\%}} / \textbf{81\%}
& 83.35 / 94\% / 84\%
& 64.25 / 78\% / \textbf{47\%} \\
\cdashline{2-8}

& Gemma 2B 
& 58.47 / 58\% / 42\% 
&52.76 / 50\% / 39\%
& 69.98 / 73\% / 62\%
&56.31 / 58\% / 47\%
& 62.17 / 62\% / 48\%
& 48.54 / 50\% / 39\% \\

& Gemma 7B 
& {70.73} / {69\%} / 55\%
& 64.67 / 62\% / 53\%
& 78.78 / 81\% / 75\%
& 67.76 / 69\% / 62\%
& 70.73 / 75\% / 61\%
& 59.77 / 59\% / 52\% \\

& Mistral 7B 
& 67.53 / 70\% / 53\%
& 61.65 / 67\% / 47\%
& 76.01 / 80\% / 72\%
& 64.43 / 70\% / 52\%
& 67.90 / 73\% / 53\%
& 54.27 / 53\% / 45\% \\

& LLaMA3 8B 
& \textcolor{black}{\textbf{81.64}} / \textcolor{black}{\textbf{88\%}} / \textbf{73\%}
& \textcolor{black}{\textbf{78.02}} / \textcolor{black}{\textbf{77\%}} / \textbf{67\%}
& \textcolor{black}{\textbf{89.20}} / \textcolor{black}{\textbf{94\%}} / \textbf{94\%}
& \textcolor{black}{\textbf{82.18}} / \textcolor{black}{\textbf{88\%}} / \textbf{75\%}
& \textcolor{black}{\textbf{83.63}} / \textcolor{black}{\textbf{94\%}} / \textbf{77\%}
& \textcolor{black}{\textbf{72.54}} / \textcolor{black}{\textbf{73\%}} / \textbf{64\%} \\
 & \cellcolor[gray]{0.93}\textbf{P-Match}$\uparrow$ 
 & \cellcolor[gray]{0.93}95.56\% / 93.62\% / 80.22\% 
 & \cellcolor[gray]{0.93}104.80\% / 95.06\% / 89.33\% 
 & \cellcolor[gray]{0.93}\textcolor{blue}{\textbf{96.03\%}} / 94\% / \textcolor{blue}{\textbf{94\%}} 
 & \cellcolor[gray]{0.93}105.41\% / \textcolor{blue}{\textbf{108.64\%}} / 92.59\% 
 & \cellcolor[gray]{0.93}95.75\% / \textcolor{blue}{\textbf{96.91\%}} / 79.38\% 
 & \cellcolor[gray]{0.93}\textcolor{blue}{\textbf{107.98\%}} / 91.25\% / \textcolor{blue}{\textbf{136.17\%}} \\
\midrule

\multirow{8}{*}{Coding}  
& GPT4o 
& \textcolor{black}{\textbf{90.31}} / \textcolor{black}{\textbf{100\%}} / \textbf{98\%}
& 75.18 / 92\% / 70\%
& \textcolor{black}{\textbf{94.57}} / \textcolor{black}{\textbf{100\%}} / \textbf{100\%}
& 86.32 / \textcolor{black}{\textbf{100\%}} / 91\%
& \textcolor{black}{\textbf{90.78}} / \textcolor{black}{\textbf{100\%}} / \textbf{100\%}
& 58.43 / 62\% / 25\% \\

& Claude 3 Sonnet 
& 88.76 / \textcolor{black}{\textbf{100\%}} / 92\%
& 75.23 / \textcolor{black}{\textbf{94\%}} / 67\%
& 93.82 / \textcolor{black}{\textbf{100\%}} / \textbf{100\%}
& \textcolor{black}{\textbf{87.42}} / \textcolor{black}{\textbf{100\%}} / \textbf{100\%}
& 89.84 / \textcolor{black}{\textbf{100\%}} / \textbf{100\%}
& 60.46 / 69\% / 31\% \\

& Gemini 1.5 Flash 
& 88.51 / 98\% / 94\%
& \textcolor{black}{\textbf{75.62}} / 91\% / \textbf{73\%}
& 93.59 / \textcolor{black}{\textbf{100\%}}  / \textbf{100\%}
& 82.92 / 97\% / 84\%
& 90.62 / \textcolor{black}{\textbf{100\%}} / 98\%
& \textcolor{black}{\textbf{64.21}} / \textcolor{black}{\textbf{84\%}} / \textbf{41\%} \\
\cdashline{2-8}

& Gemma 2B 
& 62.31 / 70\% / 44\%
& 56.48 / 66\% / 41\%
& 80.92 / 89\% / 84\%
& 67.24 / 78\% / 48\%
&  72.98 / 89\% / 66\%
& 44.08 / 50\% / 8\% \\

& Gemma 7B 
& \textcolor{black}{\textbf{80.56}} / \textcolor{black}{\textbf{92\%}} / \textbf{80\%} 
& \textcolor{black}{\textbf{71.92}} / \textcolor{black}{\textbf{89\%}} / \textbf{70\%}  
& \textcolor{black}{\textbf{90.47}} / \textcolor{black}{\textbf{100\%}} / \textbf{98\%} 
& \textcolor{black}{\textbf{80.26}} / \textcolor{black}{\textbf{92\%}} / \textbf{84\%}  
& \textcolor{black}{\textbf{84.66}} / \textcolor{black}{\textbf{100\%}} / \textbf{88\%}  
& \textcolor{black}{\textbf{61.23}} / \textcolor{black}{\textbf{72\%}} / \textbf{36\%}  \\

& Mistral 7B 
& 68.32 / 77\% / 56\%
& 61.01 / 69\% / 45\%
& 81.25 / 92\% / 81\%
& 69.10 / 83\% / 55\%
& 72.39 / 86\% / 69\% 
& 45.25 / 50\% / 8\% \\

& LLaMA3 8B 
& 77.47 / 88\% / 72\%
& 69.46 / 88\% / 61\%
& 83.97 / 94\% / 83\%
& 73.51 / 88\% / 67\%
& 75.55 / 89\% / 73\%
& 51.10 / 58\% / 17\% \\

& \cellcolor[gray]{0.93}\textbf{P-Match}$\uparrow$ 
& \cellcolor[gray]{0.93}89.20\% / 92\% / 81.63\%
& \cellcolor[gray]{0.93}95.11\% / \textcolor{blue}{\textbf{94.68\%}} / 95.89\%
& \cellcolor[gray]{0.93}\textcolor{blue}{\textbf{95.66\% / 100\%}} / \textcolor{blue}{\textbf{98\%}}
& \cellcolor[gray]{0.93}91.81\% / 92\% / 84\%
&  \cellcolor[gray]{0.93}93.26\% / \textcolor{blue}{\textbf{100\%}} / 88\%
& \cellcolor[gray]{0.93}\textcolor{blue}{\textbf{95.36\%}} / 85.71\% / \textcolor{blue}{\textbf{97.80\%}} \\
\midrule

\multirow{8}{*}{Closed QA}  
& GPT4o 
& \textcolor{black}{\textbf{95.45}} / \textcolor{black}{\textbf{100\%}} / \textbf{100\%} 
& 84.23 / 93\% / 80\%
& 97.21 / \textcolor{black}{\textbf{100\%}} / \textbf{100\%}
& 92.56 / \textcolor{black}{\textbf{100\%}} / 97\%
& 93.58 / \textcolor{black}{\textbf{100\%}} / \textbf{100\%}
& 75.58 / 85\% / 63\% \\

& Claude 3 Sonnet 
& 94.03 / \textcolor{black}{\textbf{100\%}} / 98\%
& 85.28 / \textcolor{black}{\textbf{100\%}} / 82\%
& 97.60 / \textcolor{black}{\textbf{100\%}} / \textbf{100\%}
& 93.95 / \textcolor{black}{\textbf{100\%}} / \textbf{100\%}
& 93.66 / \textcolor{black}{\textbf{100\%}} / \textbf{100\%}
& 76.33 / 92\% / 65\% \\

& Gemini 1.5 Flash 
& 94.63 / \textcolor{black}{\textbf{100\%}} / 97\%
& \textcolor{black}{\textbf{87.43}} / 95\% / \textbf{87\%}
& \textcolor{black}{\textbf{98.25}} / \textcolor{black}{\textbf{100\%}} / \textbf{100\%}
& \textcolor{black}{\textbf{97.41}} / \textcolor{black}{\textbf{100\%}} / \textbf{100\%}
& \textcolor{black}{\textbf{95.00}} / \textcolor{black}{\textbf{100\%}} / \textbf{100\%}
& \textcolor{black}{\textbf{85.66}} / \textcolor{black}{\textbf{97\%}} / \textbf{83\%} \\
\cdashline{2-8}

& Gemma 2B 
& 67.25 / 65\% / 53\%
& 67.41 / 67\% / 57\%
& 80.22 / 85\% / 78\%
& 70.20 / 73\% / 72\%
& 70.33 / 73\% / 60\%
& 59.68 / 62\% / 50\% \\

& Gemma 7B 
& 81.85 / \textcolor{black}{\textbf{88\%}} / \textbf{83\%}
& 79.02 / \textcolor{black}{\textbf{85\%}} / 78\%
& \textcolor{black}{\textbf{88.83}} / \textcolor{black}{\textbf{93\%}} / \textbf{93\%}
& 83.95 / \text{87\%} / \textbf{83\%}
& \textcolor{black}{\textbf{82.51}} / \textcolor{black}{\textbf{93\%}} / \textbf{80\%}
& \text{72.24} / \text{75\%} / 67\% \\

& Mistral 7B 
& \textcolor{black}{\textbf{83.63}} / 87\% / 82\%
& \textcolor{black}{\textbf{81.36}} / \textcolor{black}{\textbf{85\%}} / \textbf{83\%}
& 88.25 / \textcolor{black}{\textbf{93\%}} / 85\%
& \textcolor{black}{\textbf{84.77}} / \textcolor{black}{\textbf{88\%}} / \textbf{83\%}
& 82.04 / 85\% / 78\%
& \textcolor{black}{\textbf{73.95}} / \textcolor{black}{\textbf{78\%}} / \textbf{70\%} \\

& LLaMA3 8B 
& 75.55 / 78\% / 75\%
& 72.62 / 77\% / 72\%
& 86.03 / 88\% / 85\%
& 77.64 / 80\% / 80\%
& 79.09 / 85\% / 77\%
& 68.78 / 75\% / 65\% \\
& \cellcolor[gray]{0.93}\textbf{P-Match}$\uparrow$ 
& \cellcolor[gray]{0.93}87.62\% / 88\% / 83\%
& \cellcolor[gray]{0.93}\textcolor{blue}{\textbf{93.06\%}} / 85\% / \textcolor{blue}{\textbf{95.40\%}}
& \cellcolor[gray]{0.93}\textcolor{blue}{\textbf{90.41\% / 93\%}} / \textcolor{blue}{\textbf{93\%}}
& \cellcolor[gray]{0.93}87.02\% / \textcolor{blue}{\textbf{88\%}} / 83\%
& \cellcolor[gray]{0.93}86.85\% / \textcolor{blue}{\textbf{93\%}} / 80\%
& \cellcolor[gray]{0.93}86.33\% / 80.41\% / 84.34\% \\
\bottomrule
\end{tabular}
}
\label{tab:main_result}
\vspace{-2.9mm}
\end{table*}

\subsection{Implementation Details}
We implement \llamaduo using PyTorch and conduct experiments on $8\times\mathrm{A100}$ (80GB) GPUs. 

\noindent\textbf{Synthetic dataset by service LLMs.} 
We utilize the seeds selected from the train subset of the coverage dataset to prompt service LLMs to generate datasets, each comprising 300k samples. The specific prompt for data generation is presented in Figure \ref{fig:default-prompt-synthegen} of Appendix \ref{sec:prompt}.
Subsequently, we employ Locality-Sensitive Hashing (LSH) with MinHash and Rouge scoring mechanisms for data deduplication.
Specifically, the LSH MinHash can efficiently identify and remove duplicate data samples, while the Rouge scoring mechanism ensures that the curated data exhibits high-quality and meaningful variations. 
After that, we acquire 256k samples for summarization tasks and 128k for other tasks. 

\noindent\textbf{Fine-tuning Local LLMs.} 
We proceed to fine-tune the small local LLMs with $2^n\mathrm{k}, n\in\{0, 1, \ldots, 8\}$ volumes of the synthetic dataset. 
To efficiently customize local LLM for a specific downstream task within constrained environments, we leverage QLoRA \cite{dettmers2024qlora} for parameter-efficient fine-tuning with superior cost-effectiveness.
The detailed configurations, which are tailored according to dataset sizes and tasks, can be found in Appendix \ref{sec:config}.

\noindent\textbf{Batch inference.}
Each fine-tuned local model is prompted to generate $K=4$ distinct responses, with each prompt sampled from the test subsets of the coverage dataset. 
To ensure fair comparisons, we maintain a consistent batch inference configuration across all fine-tuned models.
The detailed configuration is depicted in Appendix \ref{sec:config}.

\noindent\textbf{Service LLMs as judges.}
Following \cite{zheng2024judging}, we employ pairwise comparison and single answer grading strategies to evaluate the response quality of the fine-tuned local LLMs. 
The corresponding prompts are given in Figure \ref{fig:default-prompt-assessment} of Appendix \ref{sec:prompt}. 
We utilize similarity and precision metrics. 
The similarity metric assesses the degree of correspondence between the generated responses and the ground truth, while the precision metric evaluates the accuracy of the match between the input prompts and their corresponding responses.
To ensure reliability and mitigate inherent biases in the results, both metrics are quantified on a 0 to 100 scale, with each sample undergoing evaluation $M=10$ times. 
The score of coverage percentage is set to $\zeta \in \{50, 70\}$.

\begin{figure}[t]
\centering
\includegraphics[width=\linewidth]{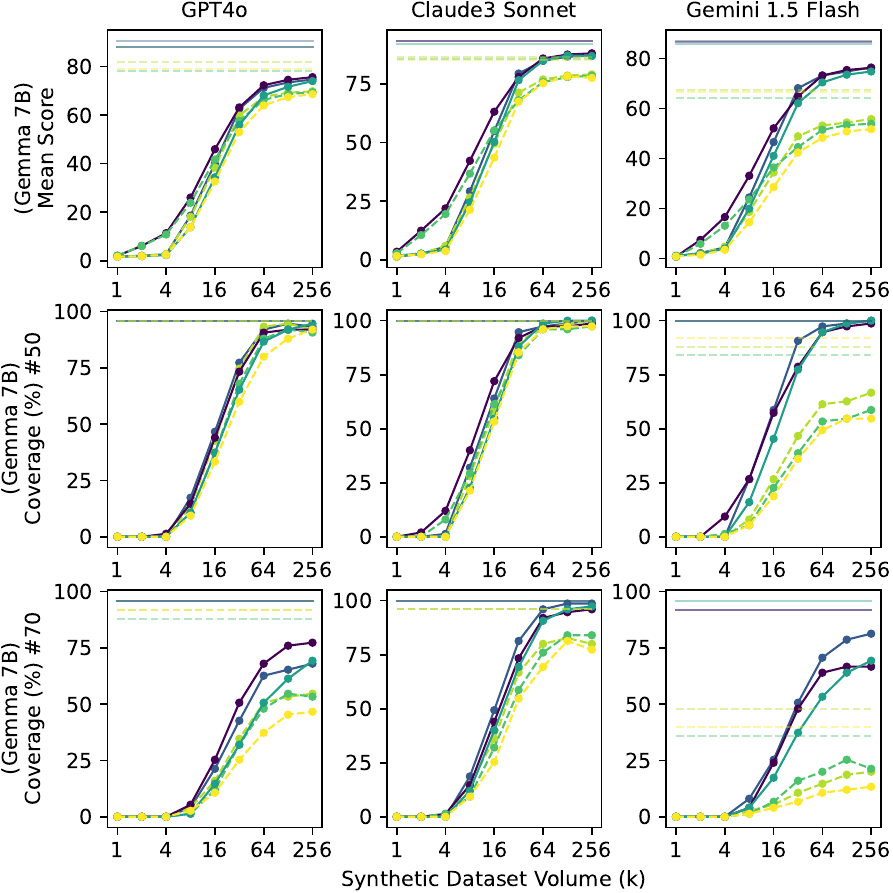}
\includegraphics[width=\linewidth]{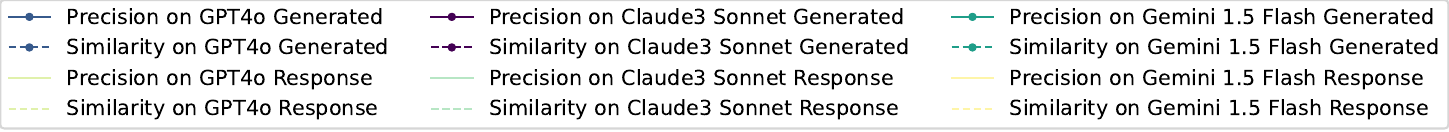}
\vspace{-3mm}
\caption{Performance of Gemma 7B fine-tuned on varied volumes of synthetic dataset produced by various service LLMs including GPT4o, Claude 3 Sonnet, and Gemini 1.5 Flash. 
The first to third columns represent the performance of the model evaluated by GPT4o, Claude 3 Sonnet, and Gemini 1.5 Flash as judges, respectively.
The first row show mean scores, while the second and third rows shows the coverage percentage with 50 and 70 scores, respectively.
}
\label{fig:gemma2b-summarization-by-three-difference-service-LLMs}
\vspace{-5mm}
\end{figure}

\begin{figure}[t]
\centering
\includegraphics[width=\linewidth]{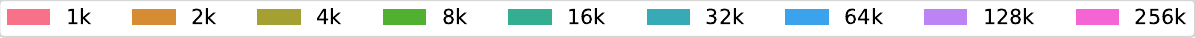}
\includegraphics[width=\linewidth]{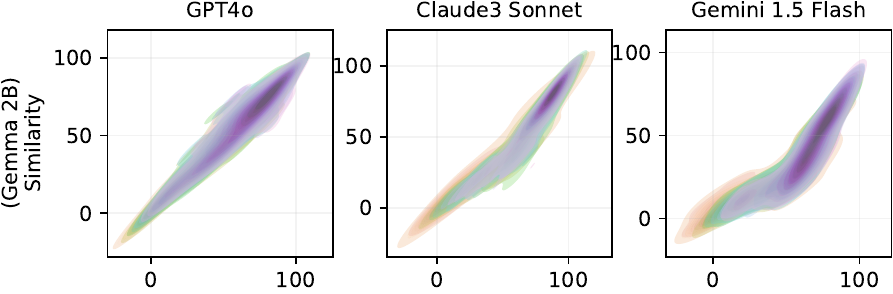}
\includegraphics[width=\linewidth]{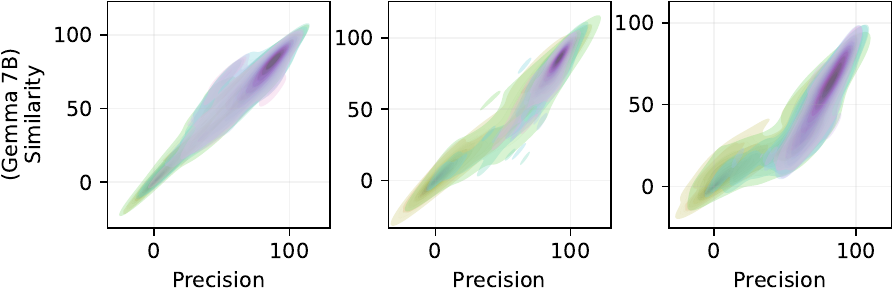}
\vspace{-3mm}
\caption{The KDE Plots of Precision v.s. Similarity by varied synthetic dataset volumes with $2^n\mathrm{k}, n\in\{0, 1, \ldots, 8\}$ and various evaluators with GPT4o, Claude 3 Sonnet, Gemini 1.5 Flash as judges from first to third columns, while the first and second rows represent the results of Gemma 2B (first row) and Gemma 7B (second row), respectively.}
\label{fig:kde-plots}
\vspace{-6mm}
\end{figure}

\subsection{Experimental Results}
This section delves into the effectiveness and adaptability of the \llamaduo pipeline, spanning different tasks with varying degrees of complexity, including summarization, classification, coding, and closed QA. 
We utilize GPT4o, Claude 3 Sonnet, and Gemini 1.5 Flash as judges to evaluate the fine-tuned model performance on test subsets of the coverage dataset.
As demonstrated in Table \ref{tab:main_result}, the fine-tuned local LLMs, despite their significantly smaller scale, achieve comparable performance on diverse tasks compared to much larger service LLMs.
For instance, in the summarization task, LLaMA3 8B achieved a comparable precision score of 87.02 / 99\% / 94\%, compared to GPT4o's score of 93.25 / 100\% / 100\%, Claude 3 Sonnet's score of 93.39 / 100\% / 99\%, and Gemini 1.5 Flash's score of 91.95 / 100\% / 98\%, with Claude 3 Sonnet serving as judge.
These results underscore the efficacy of \llamaduo in seamlessly transferring knowledge and capabilities from service LLMs to smaller local LLMs without a substantial decrease in performance.

In Table \ref{tab:main_result}, we observe distinct performance across four fine-tuned models when applied to different tasks. 
Specifically, Mistral 7B stands out in summarization tasks, achieving the best performance in 7 out of 12 cases.
Moreover, LLaMA3 8B consistently outperforms competitors across all metrics and evaluators in the classification task. 
Conversely, in coding tasks, Gemma 7B is identified as the leading model, excelling across all metrics and evaluations. 
Mistral 7B shows superior performance in the closed QA task, leading in 8 out of 12 cases. 
Within the realm of service LLMs, Claude 3 Sonnet and Gemini 1.5 Flash demonstrate exceptional performance in classification and closed QA tasks, securing the best results in 8 and 10 out of 12 cases, respectively.
Lastly, GPT4o emerges as the leading model in summarization and coding tasks, achieving the best performance in 10 and 7 out of 12 cases, respectively.
Notably, although Gemma 2B exhibits inferior performance compared to larger 7B models overall, the disparity in results is not markedly substantial, with Gemma 2B attaining closely comparable performance in certain tasks. 
For example, in closed QA tasks, Gemma 2B secures a mean precision score of 80.22, while Gemma 7B achieves 88.83, Mistral 7B reaches 88.25, and LLaMA3 8B obtains 86.03, as evaluated by Claude 3 Sonnet. 
This observation lends further support to the notion that through the strategic fine-tuning of smaller local LLMs on synthetic datasets via the \llamaduo, it is possible to closely approximate the performance of their larger counterparts.
Consequently, it offers increased flexibility and solutions for users and scenarios with budgetary considerations.
More experimental results are presented in Appendix \ref{sec:more_result}.

\subsection{In-depth LLMOps Pipeline Analysis}
In this section, we conduct an in-depth analysis of \llamaduo through summarization task. Notably, the experimental findings exhibit a consistent pattern across various tasks, underscoring the generalizability of \llamaduo.

\noindent\textbf{Impact of synthetic dataset volume.}
We explore how the volume of synthetic dataset influences the performances of fine-tuned local LLMs, aiming to elucidate a scaling law \cite{kaplan2020scaling,hoffmann2022training} on how the performance of fine-tuned models changes as the number of synthetic dataset samples increases.
Overall, the Gemma 7B model exhibits consistent performance improvements and comes closer to the performance of service LLMs with increasing volumes of synthetic data, as assessed through precision and similarity metrics by diverse evaluators, as depicted in Figure \ref{fig:gemma2b-summarization-by-three-difference-service-LLMs}.
This suggests that fine-tuning local LLMs with synthetic data, which mimics the characteristics and patterns of real-world data, can bring the same effect as actual data. Thus, it paves a new way to eliminate the challenges of data scarcity, privacy concerns, and high costs associated with crafting data \cite{liu2024best}. 
Notably, we observe that the synthetic data generated by Claude 3 Sonnet results in the highest-performing models, outperforming those fine-tuned with data produced by GPT4o and Gemini 1.5 Flash, in descending order. 
Moreover, when the synthetic dataset volume ranges from 64k to 256k, the Gemma 7B model reaches the performance saturation point and achieves performance that is much closer to, or equal to, that of service LLMs. 
This demonstrates the efficacy of our \llamaduo in enabling smaller models to replicate or even surpass the performance of service LLMs in specific downstream tasks.

\noindent\textbf{Impact of service LLMs as data generator and judge.}
As shown in Figure \ref{fig:gemma2b-summarization-by-three-difference-service-LLMs}, we observe that the choice of service LLM for data generation does not significantly impact the performance of the fine-tuned models. Specifically, (1) a consistent trend of performance enhancement is observed with the increased volume of synthetic data, irrespective of the service LLM employed for data generation; (2) the local LLMs fine-tuned on synthetic data generated by GPT4o and Claude3 Sonnet typically lead to slightly better performance than those by Gemini 1.5 Flash.
On the other hand, employing different service LLMs as judges manifests a more pronounced impact on the performance of the fine-tuned local LLMs. Overall, GPT4o and Gemini 1.5 Flash emerge as more rigorous judges compared to Claude 3 Sonnet, with Gemini 1.5 Flash assigning notably lower similarity scores. 
Moreover, we observe that in data sparsity scenarios (1k to 4k), the type of evaluators has minimal influence on the performance of the Gemma 7B model, suggesting that larger local LLMs exhibit diminished sensitivity to the choice of service LLM as a judge.
To qualitatively demonstrate the differences when using various types of service LLMs as evaluators, Figure \ref{fig:kde-plots} presents the results as KDE plots, characterized by the dataset volume. We observe that GPT4o maintains consistency in its evaluations across both similarity and precision metrics. In contrast, Claude 3 Sonnet is found to be more lenient in scoring, while Gemini 1.5 Flash assigns higher precision scores but significantly lower similarity scores. This underscores the importance of strategically aligning the selection of service LLMs with specific task requirements.

\begin{table}[!bpt]
\centering
\caption{Monthly operational cost comparison between Gemma 7B and GPT4o under different workloads. For GPT4o, input and output token counts are represented in the format input/output.}\label{tab:spec-cost-comparison}
\vspace{-2mm}
\resizebox{0.9\linewidth}{!}{
\begin{tabular}{lcccc}
\toprule
& \multicolumn{2}{c}{\text{\textbf{Light Workload}}} & \multicolumn{2}{c}{\text{\textbf{Heavy Workload}}} \\ 
\cmidrule(lr){2-3}\cmidrule(lr){4-5}
&  \text{\textbf{Gemma 7B}}  &  \text{\textbf{GPT4o}}  &  \text{\textbf{Gemma 7B}}  &  \text{\textbf{GPT4o}}  \\ 
\midrule
\multirow{2}*{\text{Fine-tuning}}     &  Cloud    &  -                       &  Cloud      &                    -     \\ 
                         &  \$800    &  -                       &  \$800      &  -                      \\ 
\midrule
\multirow{2}*{\text{Serving Specs.}}         &  1 x L4   &  300M/30M  &  8 x L4     &  1500M/150M  \\ 
                         &  \$2,539  &  \$1,950                 &  \$20,312   &  \$9,750                \\ 
\midrule
\multirow{2}*{\text{Serving Elec.}}     &  165 kWh  &  -               &  1319 kWh   &  -                      \\ 
                         &  \$30     &  -                       &  \$240      &  -                      \\ 
\midrule
\text{2 Months}        &  \text{\$3,369}  & \$3,900                  &  \text{\$21,592}   &  \$19,500               \\ 
\midrule
\text{12 Months}       &  \text{\$3,699}  & \$23,400                 &  \text{\$23,992}   &  \$117,000              \\ \bottomrule
\end{tabular}
}
\vspace{-5mm}
\end{table}

\subsection{Cost of Long-term Deployment}
\begin{figure}[!bpt]
    \centering
    \includegraphics[width=\linewidth]{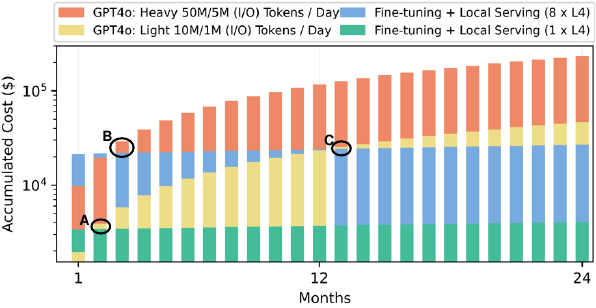}
    \vspace{-3mm}
    \caption{Long-term operational cost comparison between fine-tuning a local LLM and API-based token usage of GPT4o.  
    }
    \label{fig:cost-tradeoff}
    \vspace{-6mm}
\end{figure}

We elucidate the cost-effectiveness of our proposed \llamaduo pipeline, by conducting a long-term operational cost comparison between the fine-tuning of the small LLMs (Gemma 7B) and the token-based API usage of service LLMs (GPT4o).
In the context of local LLM deployment, the QLoRA fine-tuning process of Gemma 7B, utilizing a dataset containing 256K samples, necessitates approximately one hour to complete a single experiment on $8\times\mathrm{A100}$ GPUs.
This process incurs an estimated cost of \$50, based on the price provided by Google Cloud Platform. 
Accounting for multiple iterations of hyperparameter optimization, we estimate that the total fine-tuning cost remains below \$800, which is deemed to be negligible.
Deploying a single instance of the Gemma 7B model with support for a 1024 context length necessitates 24GB of GPU memory, making the L4 GPU an appropriate choice. Depending on the projected workload, the Gemma 7B model can be deployed either on a single server equipped with one L4 GPU (\$2,539) or across eight servers, each with one L4 GPU, with each server hosting a replica of the model instance (\$20,312). In addition, the power consumption for each server is approximately \$30 per month.
For GPT4o, as of August 2024, the pricing is \$5 and \$15 per million tokens for input and output, respectively. We estimate that a light workload, utilizing 10 million input tokens and 1 million output tokens per day, incurs a daily cost of \$65. Conversely, a heavy workload, consuming 50 million input tokens and 10 million output tokens per day, is estimated to cost \$325 daily.
The monthly operational cost comparison between Gemma 7B and GPT4o under different workloads is summarized in Table \ref{tab:spec-cost-comparison}, demonstrating a significant advantage in fine-tuning and deploying a local LLM. 
Moreover, as depicted in Figure \ref{fig:cost-tradeoff}, after the first two months, the cost of using GPT4o under both light and heavy workloads exceeds that of setting up and running a local model deployed on $1\times\mathrm{L4}$ GPU and $8\times\mathrm{L4}$ GPU, respectively, as indicated by markers \textbf{A} and \textbf{B}. 
After one year, GPT4o's costs surpass those of deploying a local model in all scenarios, as denoted by marker \textbf{C}.
These findings highlight the substantial economic benefits of investing in local LLM fine-tuning and deployment for long-term use. 
Avoiding recurring token-based charges and maintaining control over model customization further enhances the appeal of the \llamaduo for cost-conscious users and scenarios.

\section{Conclusion}
In this study, we introduce \llamaduo, the first automatic LLMOps pipeline designed to facilitate the seamless migration from service-oriented LLMs to smaller, locally manageable models.
We conduct extensive experiments and analysis across a range of tasks with popular service and local LLMs to substantiate that \llamaduo guarantees smaller local LLMs possess the potential to match or even exceed the performance of service LLMs in specific downstream tasks, providing a promising research direction to maintain cloud-based LLMs' service continuity in constrained environments. 

\section*{Limitations}
While our \llamaduo pipeline presents a promising solution for migrating capabilities from service-oriented LLMs to smaller local models, as depicted in Table \ref{tab:main_result}, several limitations must be acknowledged. 
First, the reliance on synthetic datasets generated by the service LLM may introduce biases and safety issues inherent in the original model, potentially affecting the fine-tuned model's performance on specific tasks or datasets \cite{liu2024best}. 
Additionally, the effectiveness of the pipeline in transferring knowledge is contingent upon the quality and diversity of the synthetic data generated. If the data does not adequately cover the necessary scope, the fine-tuned model may struggle with tasks outside of the provided examples \cite{razeghi2022impact,kandpal2023large}. 
Furthermore, the iterative fine-tuning process, while beneficial for performance enhancement, can be computationally intensive and time-consuming, potentially offsetting some gains in model efficiency, cost, and affordability. Another limitation is the potential plateau in performance gains after several SFT iterations, which could necessitate alternative strategies for further improvement, \textit{e.g.,} reinforcement learning (RL) \cite{ouyang2022training,rafailov2023direct}. 
Lastly, the pipeline assumes access to the service LLM for data generation, which may not always be feasible due to proprietary restrictions or API access limitations.

\section*{Ethical Considerations}
Our work introduces several ethical considerations that require careful examination. Primarily, the process of generating synthetic datasets raises questions about data privacy and security, especially if the data contains sensitive or proprietary information. There is a risk that such data, if not properly anonymized and secured, could lead to privacy violations or unauthorized data exposure \cite{liu2024best,das2025security}. 
Moreover, the transfer of biases from the service LLM to the smaller model could perpetuate or even exacerbate existing biases, leading to unfair or discriminatory outcomes in certain applications. It is crucial to implement robust bias detection and mitigation strategies within the pipeline to safeguard against these risks. Additionally, the use of proprietary models for generating synthetic data necessitates transparency regarding data handling practices and the potential limitations of the resultant models \cite{wang2023self}.

\section*{Broader Impact}
Beyond the immediate focus of this paper, we believe that the introduction of the \llamaduo pipeline has the potential to significantly impact the landscape of LLMs deployment, particularly in environments with constrained resources or stringent privacy requirements. 
By enabling the migration of capabilities from large service-oriented LLMs to smaller, locally manageable models, the pipeline empowers organizations to maintain LLMs functionalities independently of external service providers, enhancing operational resilience and reducing dependency. This can lead to increased accessibility to advanced LLMs for smaller entities or those operating in regions with limited internet connectivity. 

\section*{Acknowledgements}
Jing Tang's work is partially supported by National Key R\&D Program of China under Grant No.\ 2024YFA1012700 and No.\ 2023YFF0725100, by the National Natural Science Foundation of China (NSFC) under Grant No.\ 62402410 and No.\ U22B2060, by Guangdong Provincial Project (No.\ 2023QN10X025), by Guangdong Basic and Applied Basic Research Foundation under Grant No.\ 2023A1515110131, by Guangzhou Municipal Science and Technology Bureau under Grant No.\ 2024A04J4454, by Guangzhou Municipal Education Bureau (No.\ 2024312263), by Nansha District Project (No.\ 2023ZD022), and by Guangzhou Industrial Information and Intelligent Key Laboratory Project (No.\ 2024A03J0628) and Guangzhou Municipal Key Laboratory of Financial Technology Cutting-Edge Research (No.\ 2024A03J0630). This work is supported by IITP grant funded by the Korea government(MSIT)[RS-2023-00215959, Development of Access Agnostic wired and wire-less integrated optical access technology].

\bibliography{custom}

\clearpage
\appendix

\section{Prompt Templates}
\label{sec:prompt}
In the \llamaduo pipeline, we employ two prompt templates that serve different purposes: one for the generation of synthetic datasets and another for the evaluation of the outputs produced by the fine-tuned LLMs.

\begin{figure}[ht!]
\centering
\includegraphics[width=\linewidth]{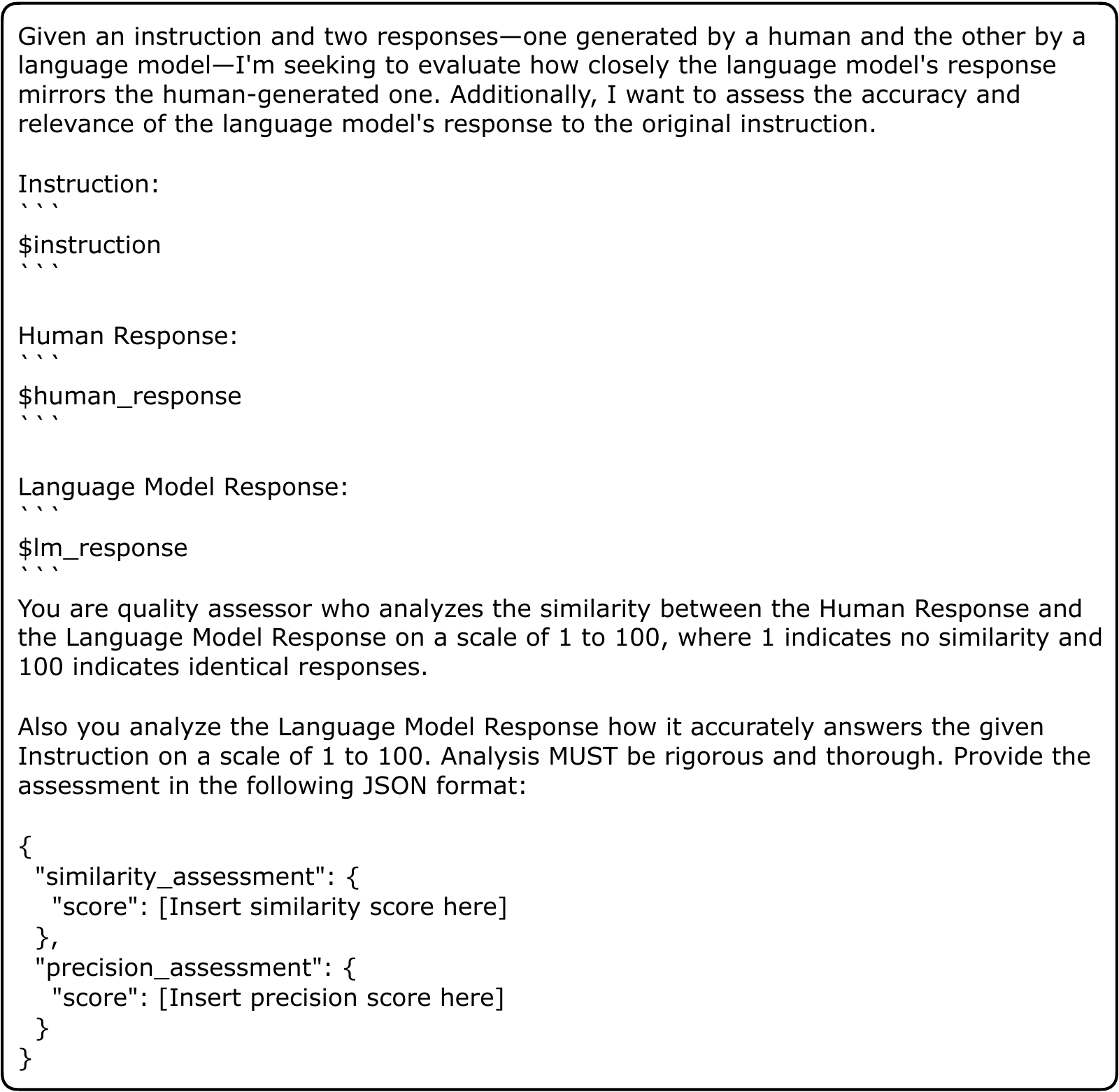}
\caption{Prompt template to evaluate the fine-tuned model's response.}
\label{fig:default-prompt-assessment}
\end{figure}

Figure \ref{fig:default-prompt-assessment} illustrates the prompt template used to assess the precision and similarity of the response \texttt{\$lm\_response} generated by fine-tuned small-scale LLMs, based on the prompt \texttt{\$instruction} and response \texttt{\$human\_response} from the test subset of the coverage dataset. It is important to note that the \texttt{\$} symbol indicates a placeholder, designed to be substituted with actual data during the runtime.

\begin{figure}[t!]
\centering
\includegraphics[width=\linewidth]{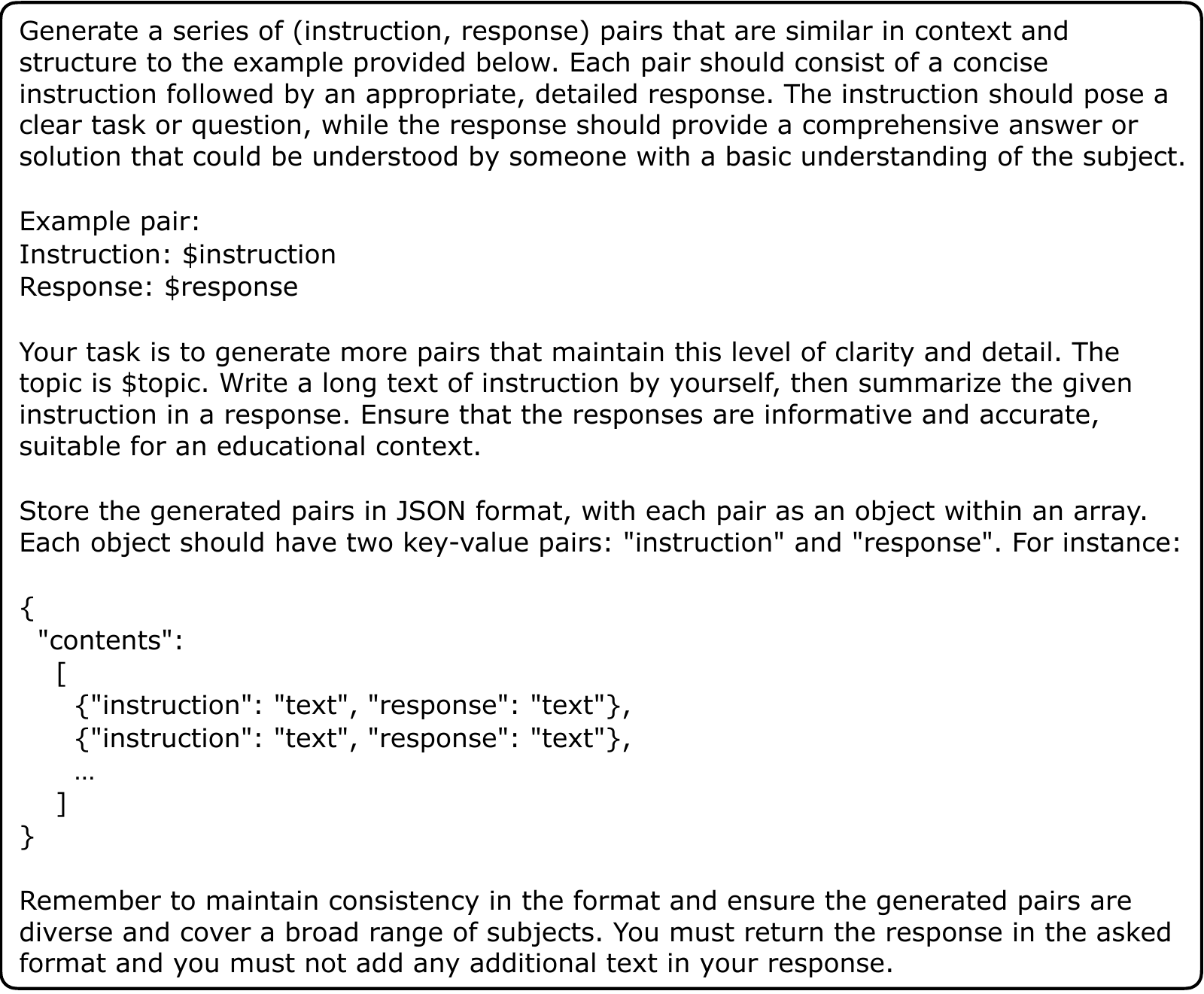}
\caption{Prompt template of data synthesis for summarization tasks.}
\label{fig:default-prompt-synthegen}
\end{figure}

\begin{figure}[t!]
\centering
\includegraphics[width=\linewidth]{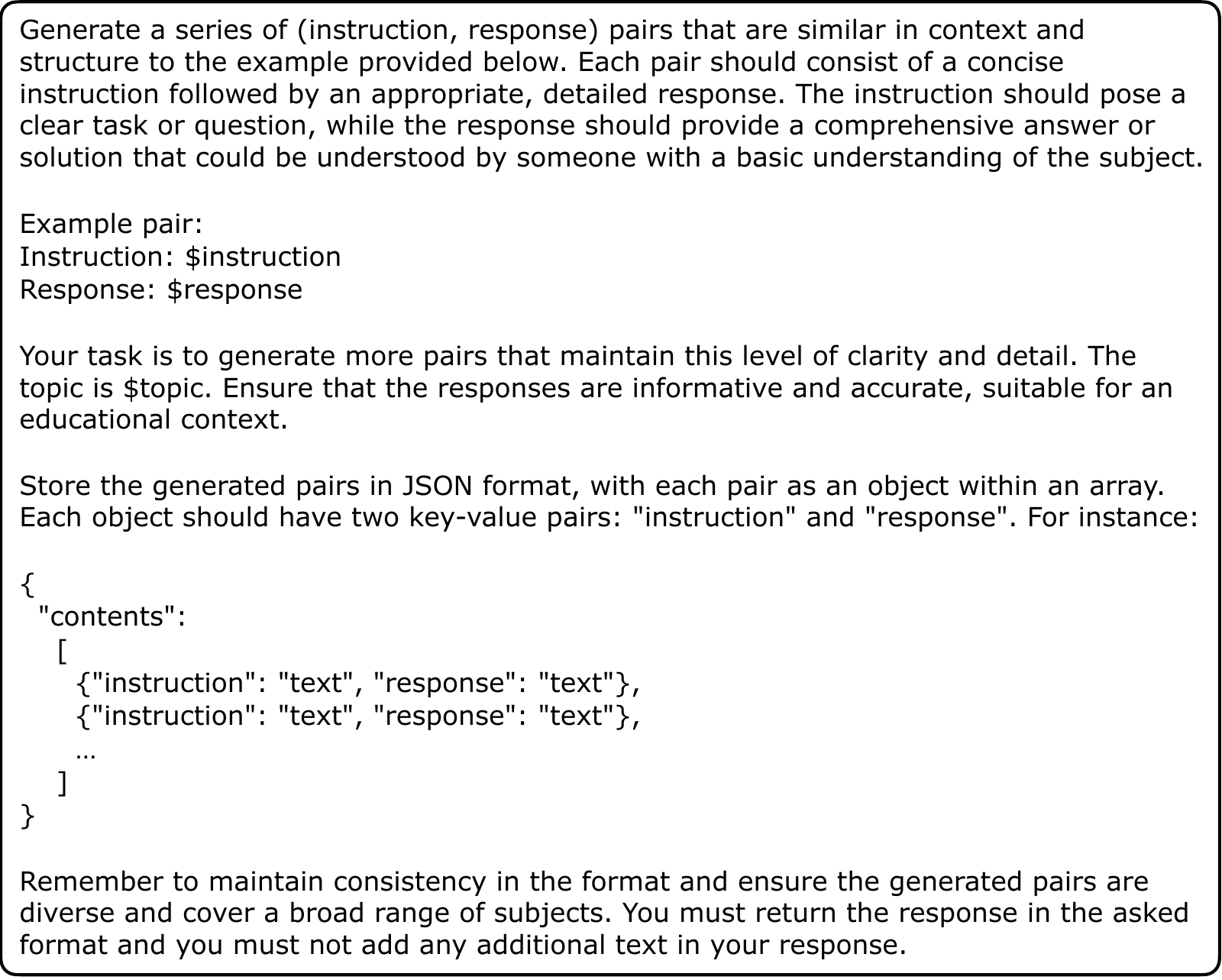}
\caption{Prompt template of data synthesis for classification, coding, and closed QA tasks.}
\label{fig:default-prompt-synthegen2}
\end{figure}

Figure \ref{fig:default-prompt-synthegen} shows the prompt template designed for the generation of synthetic data tailored to the summarization task while Figure \ref{fig:default-prompt-synthegen2} shows the prompt template for other tasks. Specifically, we use a prompt \texttt{\$instruction} alongside its corresponding response \texttt{\$response}, both sourced from the train subset of the coverage dataset, serving as an example pair.
This example pair is utilized to instruct service LLMs to generate analogous data samples. 
In addition, our template is designed to generate multiple synthetic data samples through a singular request, thereby enhancing the efficiency of API utilization.
Due to the unique features of different downstream tasks, there is no optimal prompt template that universally applies. The actual content of the prompt template is adjusted to align with the specific requirements of the task for which the synthetic dataset is being generated.

\section{Implementation Configuration}\label{sec:config}
This section delineates the statistical information of the coverage dataset and synthetic dataset generated by service LLMs. In addition, we present the details of the training configurations of our experiments.
The detailed pipeline implementation of \llamaduo is available at \textcolor{blue}{\url{https://github.com/deep-diver/llamaduo}}.

\subsection{Coverage Datasets}
The foundational coverage dataset employed in our study is the ``No Robots'' dataset \cite{no_robots}. 
We leverage four subsets of the coverage dataset, namely summarization, classification, coding, and closed QA, for synthetic data generation.
Table \ref{tab:coverage_dataset_volume} illustrates the initial composition of the task-specific subsets, with each initially containing approximately 300 original data points. 
These subsets are subsequently expanded to encompass more data points using the \llamaduo framework.
To perform an in-depth analysis of the behavior of different service LLMs, we create synthetic datasets for the summarization task by utilizing GPT4o, Claude 3 Sonnet, and Gemini 1.5 Flash. For all other tasks, we exclusively use GPT4o, owing to budget constraints.

\begin{table}[t]
\caption{Volume of coverage dataset before and after \llamaduo pipeline.}
\label{tab:coverage_dataset_volume}
\centering
\resizebox{\linewidth}{!}{ 
\begin{tabular}{l|ccc} 
\text{\textbf{Task}} & \text{\textbf{Split}} & \text{\textbf{Before}} & \text{\textbf{After}} \\ 
\midrule 
\multirow{2}{*}{\text{Summarization}\newline\text{(GPT4o)}} & train & 395 & 256K \\
                                                  & test  & 25  & 100  \\
\multirow{2}{*}{\text{Summarization}\newline\text{(Claude 3 Sonnet)}} & train & 395 & 256K \\
                                                  & test  & 25  & 100  \\
\multirow{2}{*}{\text{Summarization}\newline\text{(Gemini 1.5 Flash)}} & train & 395 & 256K \\
                                                  & test  & 25  & 100  \\
\midrule 
\multirow{2}*{Classification(GPT4o)} & train & 334 & 128K \\
                       & test  & 16  & 64  \\
\midrule 
\multirow{2}*{Coding(GPT4o)}        & train & 334 & 128K \\
                       & test  & 16  & 64  \\
\midrule 
\multirow{2}*{Closed QA(GPT4o)}     & train & 245 & 128K \\
                       & test  & 15  & 60  \\
\end{tabular}%
}
\end{table}

\begin{table}[!bpt]
\caption{Token-level statistics of the coverage and synthetic datasets.}
\label{tab:coverage_dataset_tokens}
\centering
\resizebox{\linewidth}{!}{%
\begin{tabular}{l|cccc}
\text{\textbf{Task}} & \text{\textbf{Min}} & \text{\textbf{Max}} & \text{\textbf{Avg.}} & \text{\textbf{Std.}} \\
\midrule 
\text{Summarization (Coverage-Train)} & 85 & 2386 & 389 & 256 \\
\text{Summarization (Coverage-Test)} & 148 & 1150 & 426 & 245 \\
\text{Summarization (GPT4o)} & 10 & 2386 & 95 & 53 \\
\text{Summarization (Claude 3 Sonnet)} & 10 & 2386 & 118 & 64 \\
\text{Summarization (Gemini 1.5 Flash)} & 10 & 2386 & 108 & 62 \\
\midrule 
\text{Classification (Coverage-Train)} & 18 & 2159 & 207 & 244 \\
\text{Classification (Coverage-Test)} & 46 & 520 & 119 & 109 \\
\text{Classification (GPT4o)} & 6 & 2159 & 67 & 37 \\
\midrule 
\text{Coding (Coverage-Train)} & 38 & 6518 & 350 & 502 \\
\text{Coding (Coverage-Test)} & 49 & 821 & 317 & 189 \\
\text{Coding (GPT4o)} & 9 & 6518 & 151 & 84 \\
\midrule 
\text{Closed QA (Coverage-Train)} & 58 & 1497 & 320 & 241 \\
\text{Closed QA (Coverage-Test)} & 126 & 1578 & 411 & 378 \\
\text{Closed QA (GPT4o)} & 12 & 1701 & 135 & 59
\end{tabular}%
}
\end{table}

Table \ref{tab:coverage_dataset_tokens} presents the statistical information of the token count across each dataset. 
We only use data from the coverage train set for data synthesis and alignment tasks.
We observe a reduction in both the average number of tokens and the standard deviation across the synthetic datasets compared to the original dataset. This is due to that the data synthesis process generates multiple synthetic data samples within a single API request.

\subsection{Training Configurations}
We utilize Hugging Face’s ``Alignment Handbook'' \cite{alignment_handbook2023} and the alignment recipes tailored for the Gemma models to streamline the fine-tuning process.

\begin{table}[t]
\caption{Detailed configurations used in the experiments.}
\label{tab:training-configurations}
\centering
\resizebox{0.9\linewidth}{!}{%
\begin{tabular}{c|ll}
\multicolumn{1}{l|}{} & \multicolumn{1}{l}{\text{\textbf{Configuration}}} & \multicolumn{1}{c}{\text{\textbf{Value}}} \\
\midrule 
\multirow{6}{*}{\text{Common}} & Data Type & bfloat16 \\
 & Learning Rate Scheduler & cosine \\
 & Max Number of Tokens & 1024 \\
 & LoRA Type & QLoRA \\
 & LoRA Dropout & 0.05 \\
\midrule 
\multirow{2}{*}{\text{1K$\sim$16K}} & LoRA Rank & 8 \\
 & LoRA Alpha & 16 \\
\midrule 
\multirow{2}{*}{\text{32K}} & LoRA Rank & 16 \\
 & LoRA Alpha & 32 \\
\midrule 
\multirow{2}{*}{\text{64K$\sim$256K}} & LoRA Rank & 32 \\
 & LoRA Alpha & 64
\end{tabular}%
}
\end{table}

\begin{figure*}[t]
\centering
\includegraphics[width=0.71\linewidth]{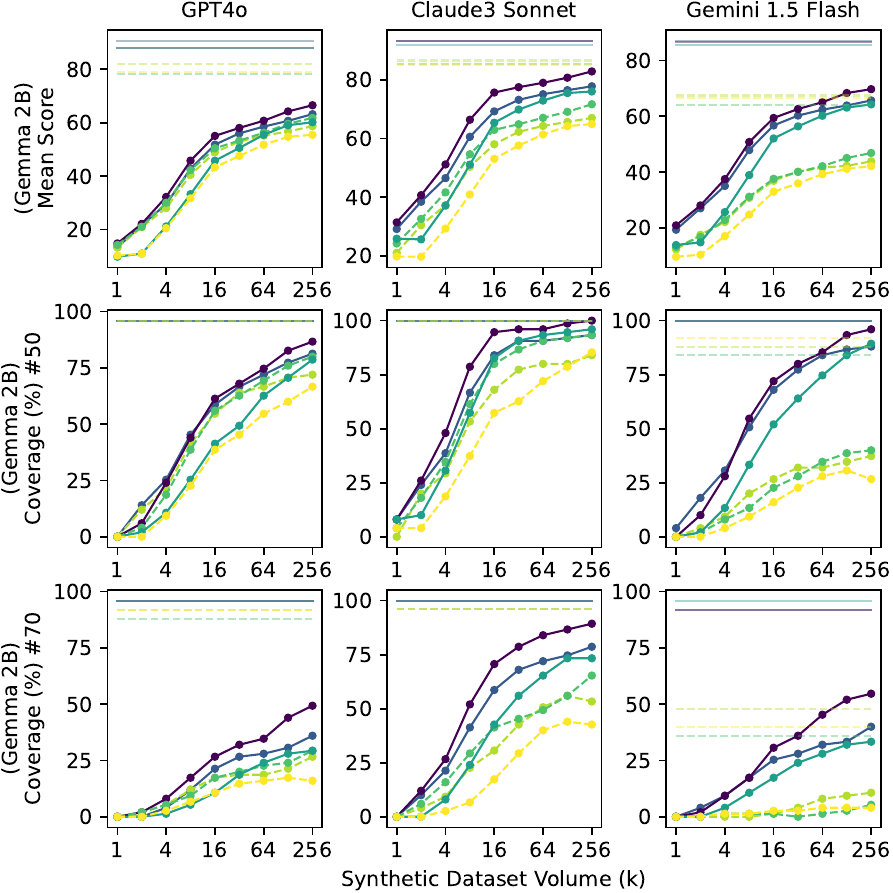}
\includegraphics[width=0.7\linewidth]{images/legend_full.pdf}
\caption{Performance of Gemma 2B fine-tuned on varied volumes of synthetic dataset producted by various service LLMs including GPT4o, Claude 3 Sonnet, and Gemini 1.5 Flash. 
The first to third columns represent the performance of the model evaluated by GPT4o, Claude 3 Sonnet, and Gemini 1.5 Flash as judges, respectively.
The first row show mean scores, while the second and third rows show the coverage percentage with 50 and 70 scores, respectively. 
}
\label{fig:gemma2b-summarization-by-three-difference-service-LLMs-full}
\end{figure*}

As outlined in Table \ref{tab:training-configurations}, we employ QLoRA \cite{dettmers2024qlora} to align the Gemma 2B and 7B, Mistral 7B, and LLaMA3 8B models efficiently. The QLoRA method leverages the advantages of low-rank adaptation, reducing the computational resources required for training. Throughout the alignment procedure, we incrementally adjust the rank and alpha values of LoRA, aiming to optimize the adaptation layer’s capacity to match the increasing complexity of the datasets.

We set the maximum token as 1024 for the training phase, notwithstanding the presence of data samples exceeding this threshold.
This decision is made based on a comprehensive analysis of the dataset, which indicates that data samples surpassing the token limit constitute a negligible portion of the total dataset.
By imposing this limitation, we can concentrate our computational efforts on the majority of the data, thereby enhancing the efficiency of training without significantly compromising the models’ ability to generalize to real-world scenarios.

The 1024-token limit, though seemingly restrictive, does not impede the performance of the aligned fine-tuned small-scale models. 
All fine-tuned models exhibit robust performances across the experiments, as they are trained and evaluated on data predominantly falling within the 1024-token boundary.
This outcome corroborates our analysis of the data and demonstrates the efficacy of QLoRA, even within the constraints of our allocated computational budget.

\section{More Experimental Results}\label{sec:more_result} 
The performance of Gemma 2B fine-tuned on varied volumes of synthetic dataset produced by various service LLMs including GPT4o, Claude 3 Sonnet, and Gemini 1.5 Flash is shown in Figure \ref{fig:gemma2b-summarization-by-three-difference-service-LLMs-full}.

\section{Case Study}\label{sec:case_study}
This section delves into detailed case studies showcasing the enhanced capabilities of the aligned small-scale local LLMs. 
We use Gemma 2B and 7B models as examples to illustrate.

\begin{figure*}[htbp]
\begin{minipage}{\textwidth}
\centering
\includegraphics[width=0.9\linewidth]{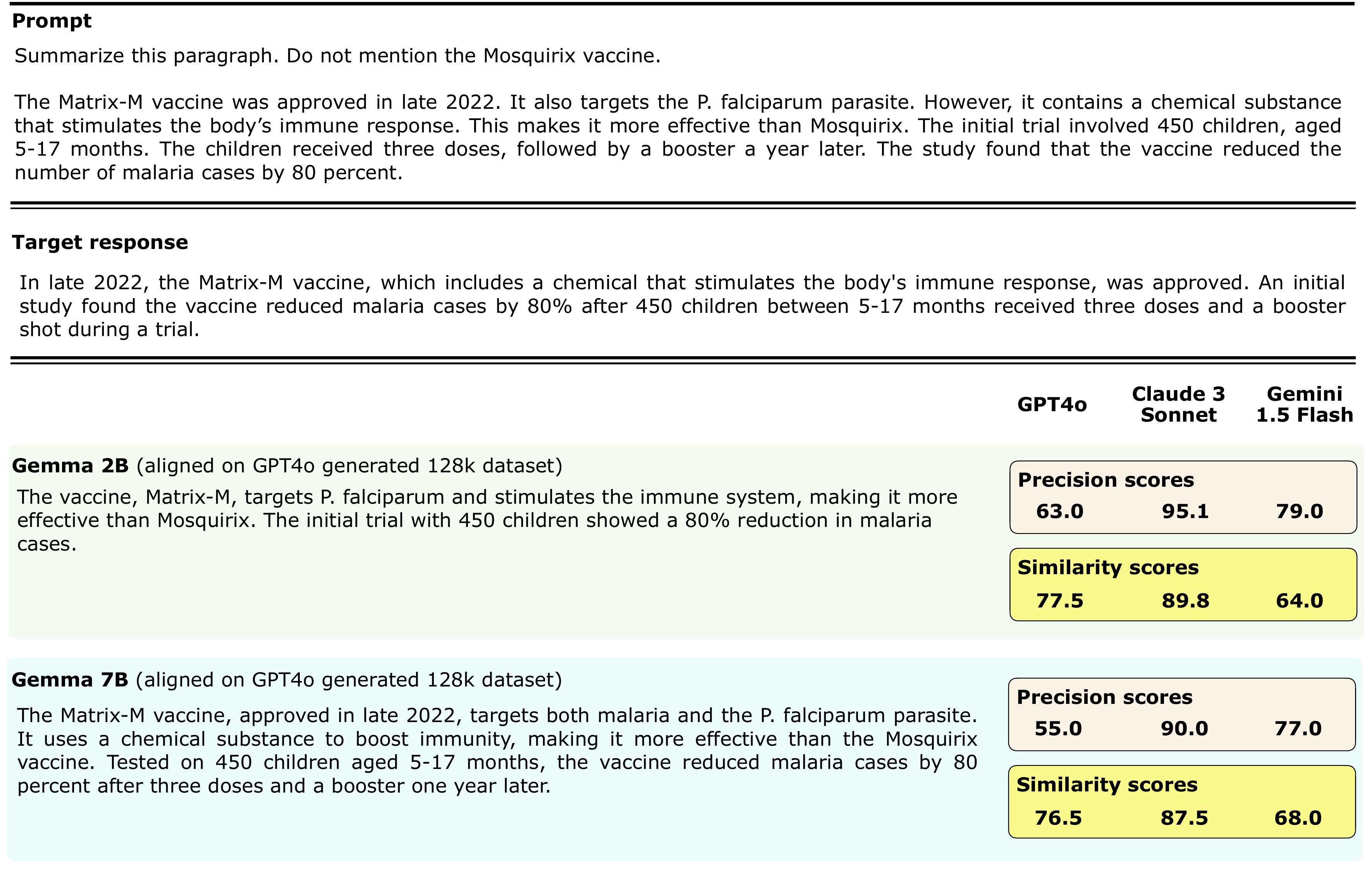}
\caption{Responses by Gemma 2B and Gemma 7B models fine-tuned on GPT4o generated 128K synthetic dataset for summarization task.}
\label{fig:example1}
\end{minipage}
\begin{minipage}{\textwidth}
\centering
\includegraphics[width=0.9\linewidth]{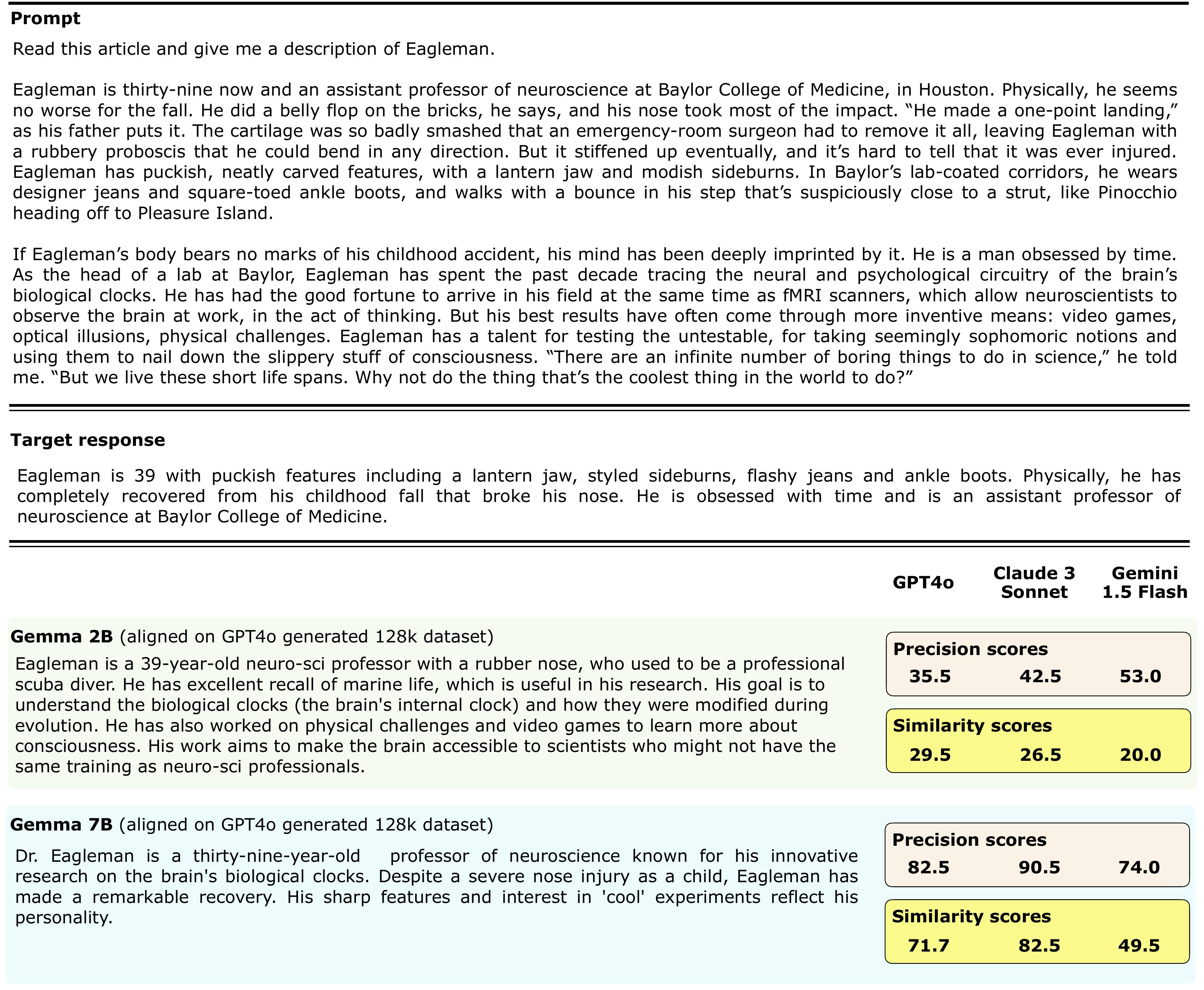}
\caption{Responses by Gemma 2B and Gemma 7B models fine-tuned on GPT4o generated 128K synthetic dataset for summarization task.}
\label{fig:example1-2}    
\end{minipage}
\end{figure*}

\begin{figure*}[ht!]
\centering
\includegraphics[width=0.9\linewidth]{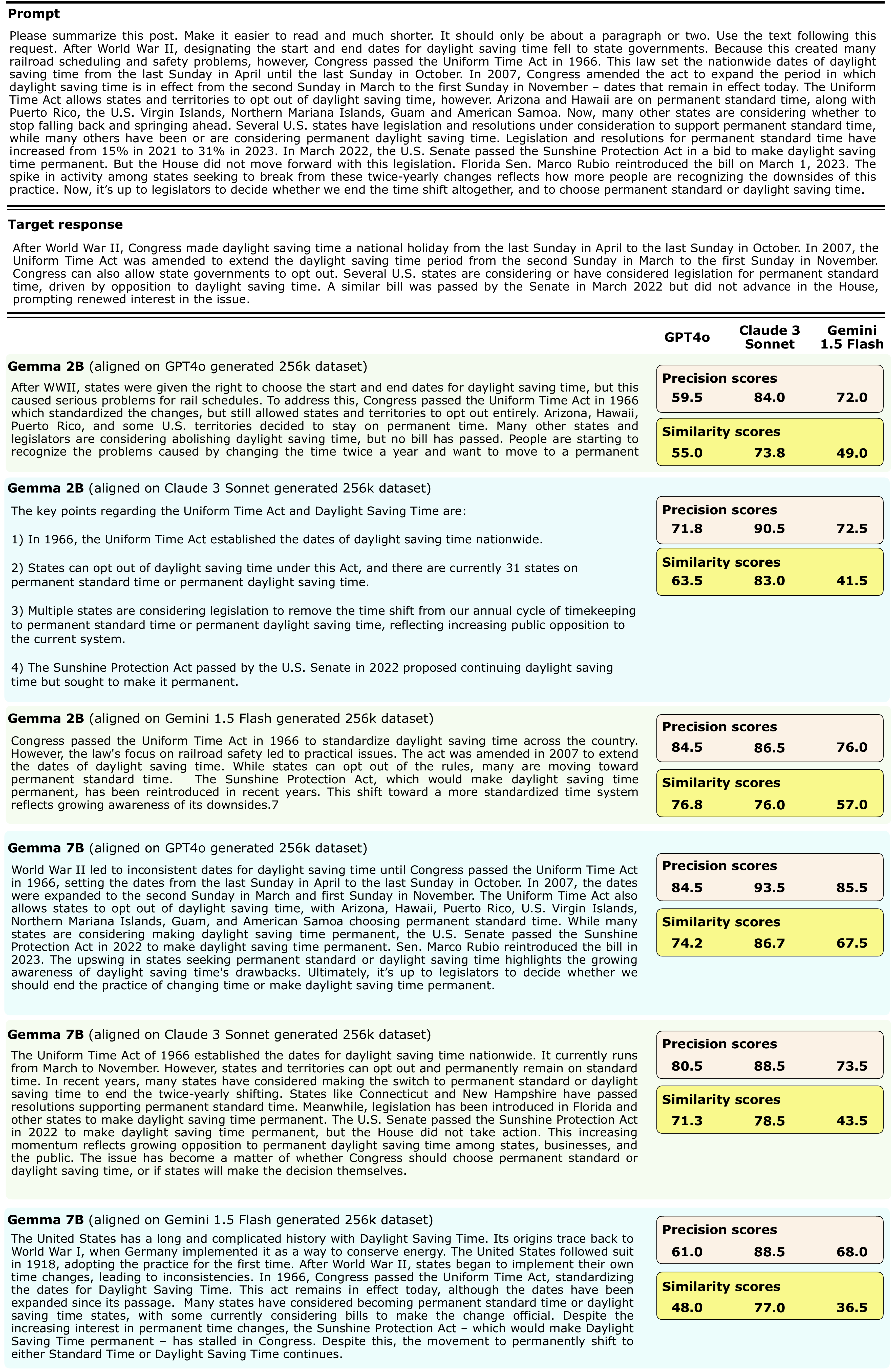}
\caption{Responses by Gemma 2B and Gemma 7B models fine-tuned on {GPT4o, Claude 3 Sonnet, Gemini 1.5 Flash} generated 256K of synthetic datasets for summarization task.}
\label{fig:example5}
\end{figure*}

\begin{figure*}[ht!]
\centering
\includegraphics[width=0.95\linewidth]{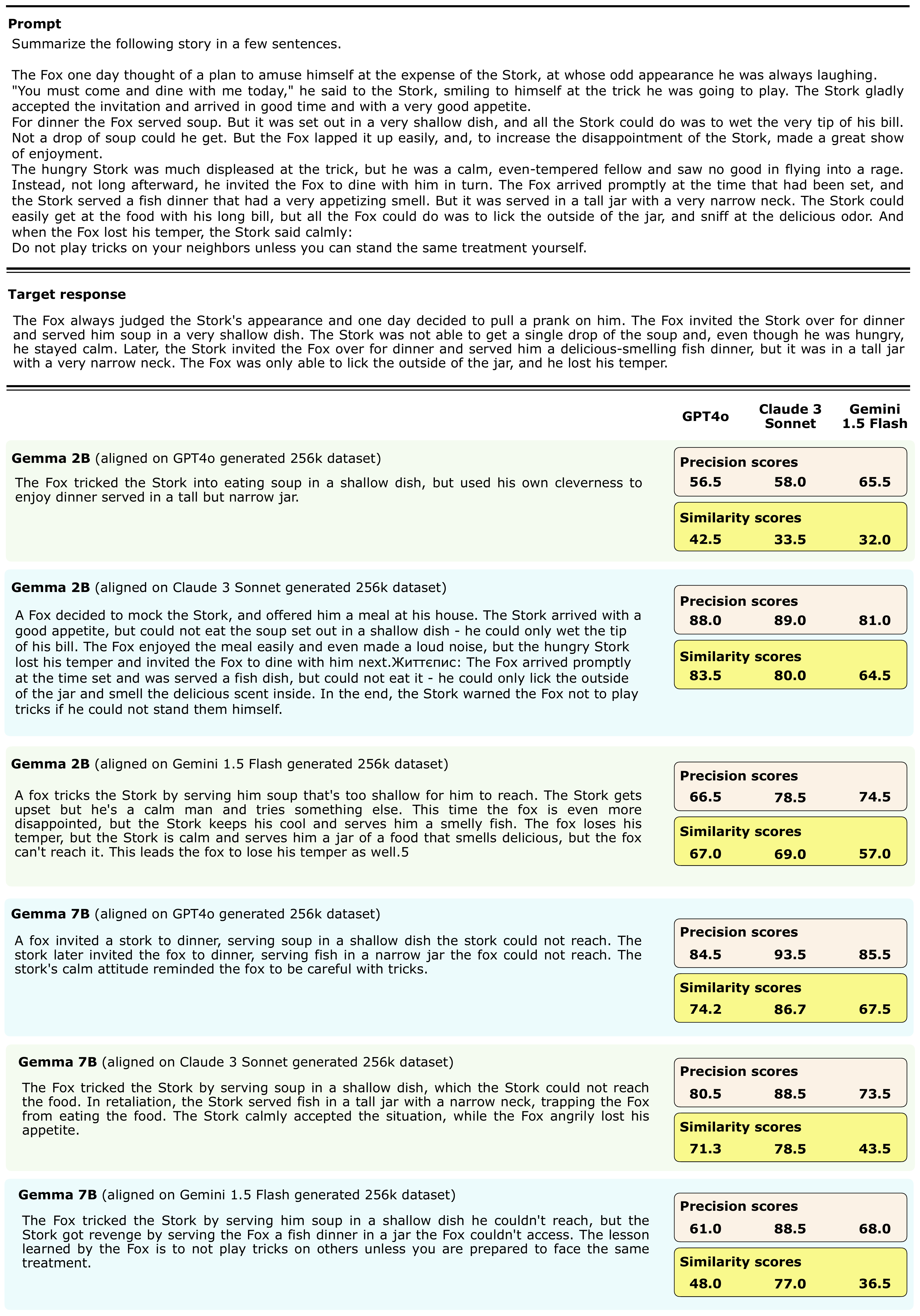}
\caption{Responses by Gemma 2B and Gemma 7B models fine-tuned on {GPT4o, Claude 3 Sonnet, Gemini 1.5 Flash} generated 256K of synthetic datasets for summarization task.}
\label{fig:example6}
\end{figure*}

\begin{figure*}[htbp]
\begin{minipage}{\textwidth}
\centering
\includegraphics[width=0.9\linewidth]{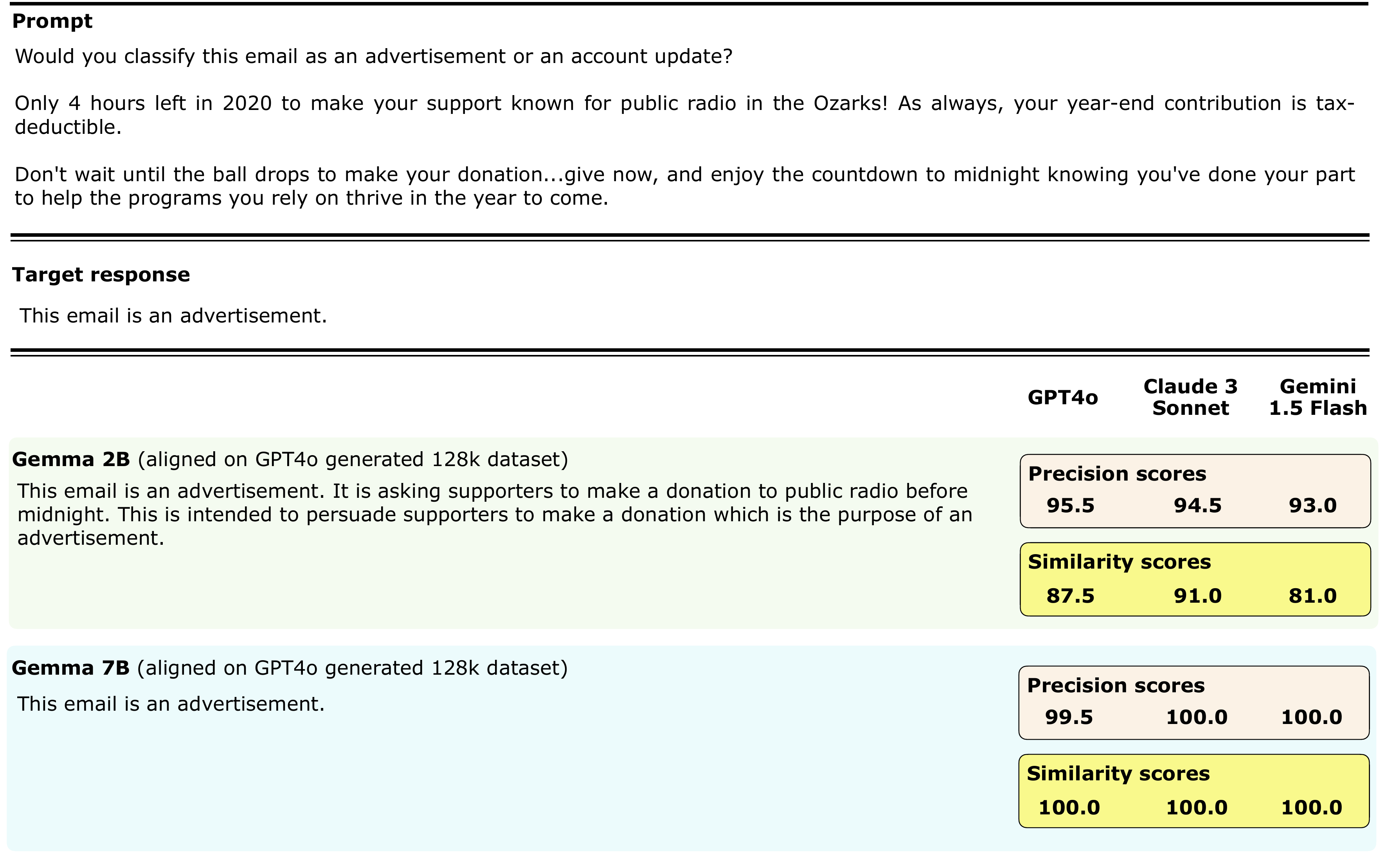}
\caption{Responses by Gemma 2B and Gemma 7B models fine-tuned on GPT4o generated 128K synthetic dataset for classification task.}
\label{fig:example2}
\end{minipage}

\begin{minipage}{\textwidth}
\centering
\includegraphics[width=0.9\linewidth]{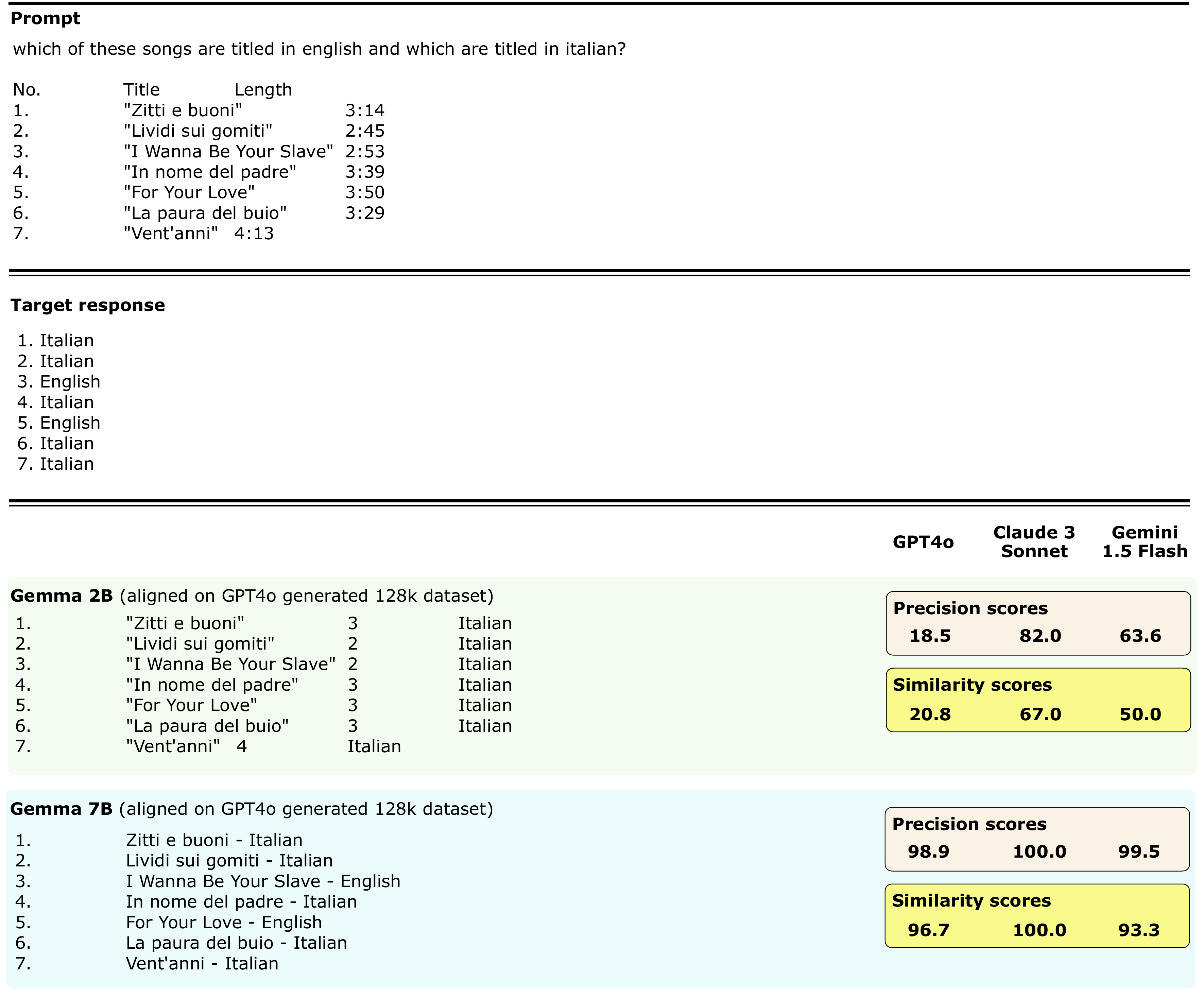}
\caption{Responses by Gemma 2B and Gemma 7B models fine-tuned on GPT4o generated 128K synthetic dataset for classification task.}
\label{fig:example2-2}
\end{minipage}
\end{figure*}

\begin{figure*}[ht!]
\centering
\includegraphics[width=\linewidth]{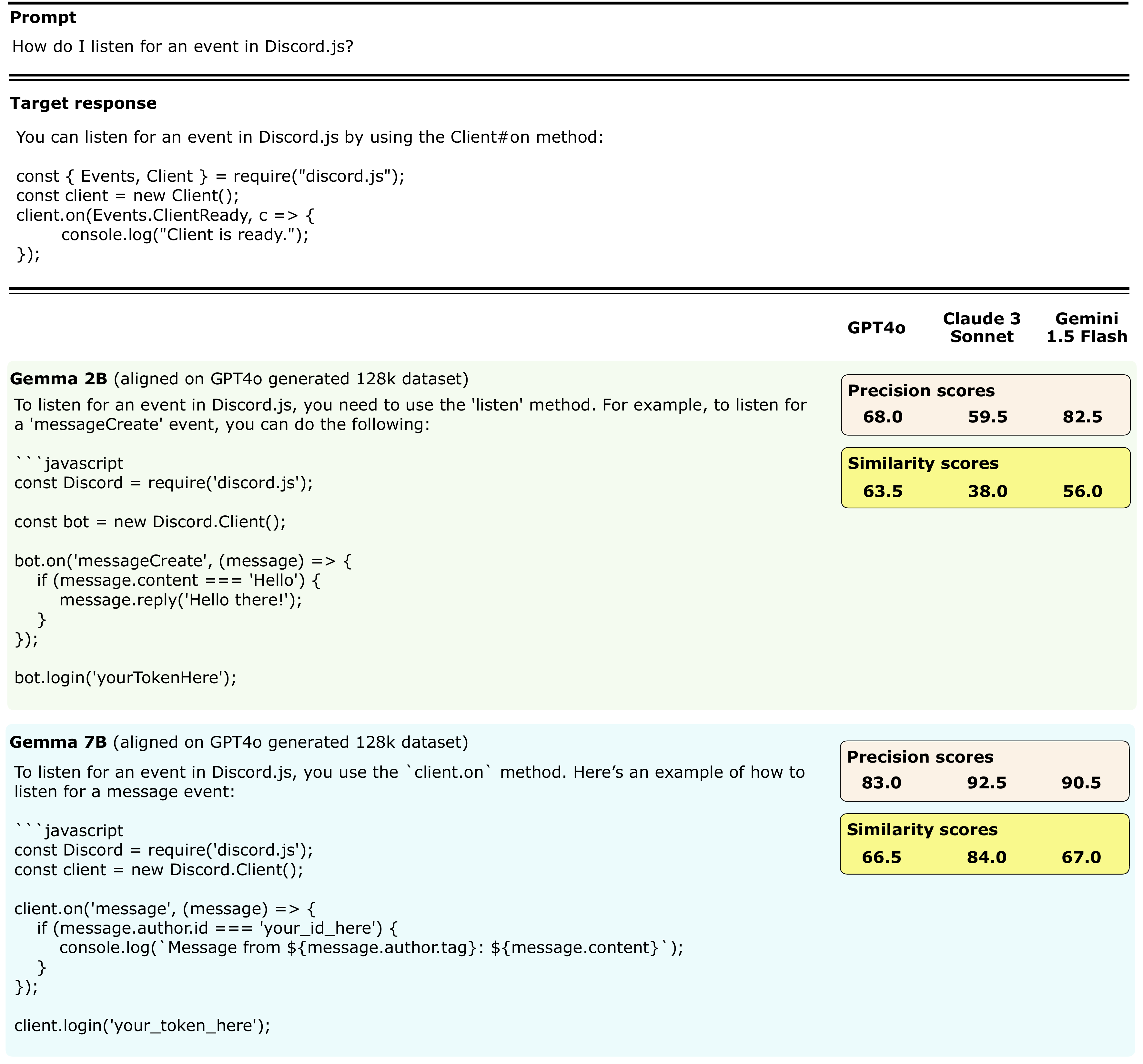}
\caption{Responses by Gemma 2B and Gemma 7B models fine-tuned on GPT4o generated 128K synthetic dataset for coding task.}
\label{fig:example3}
\end{figure*}

\begin{figure*}[ht!]
\centering
\begin{minipage}[b]{0.95\linewidth} 
\includegraphics[width=\linewidth]{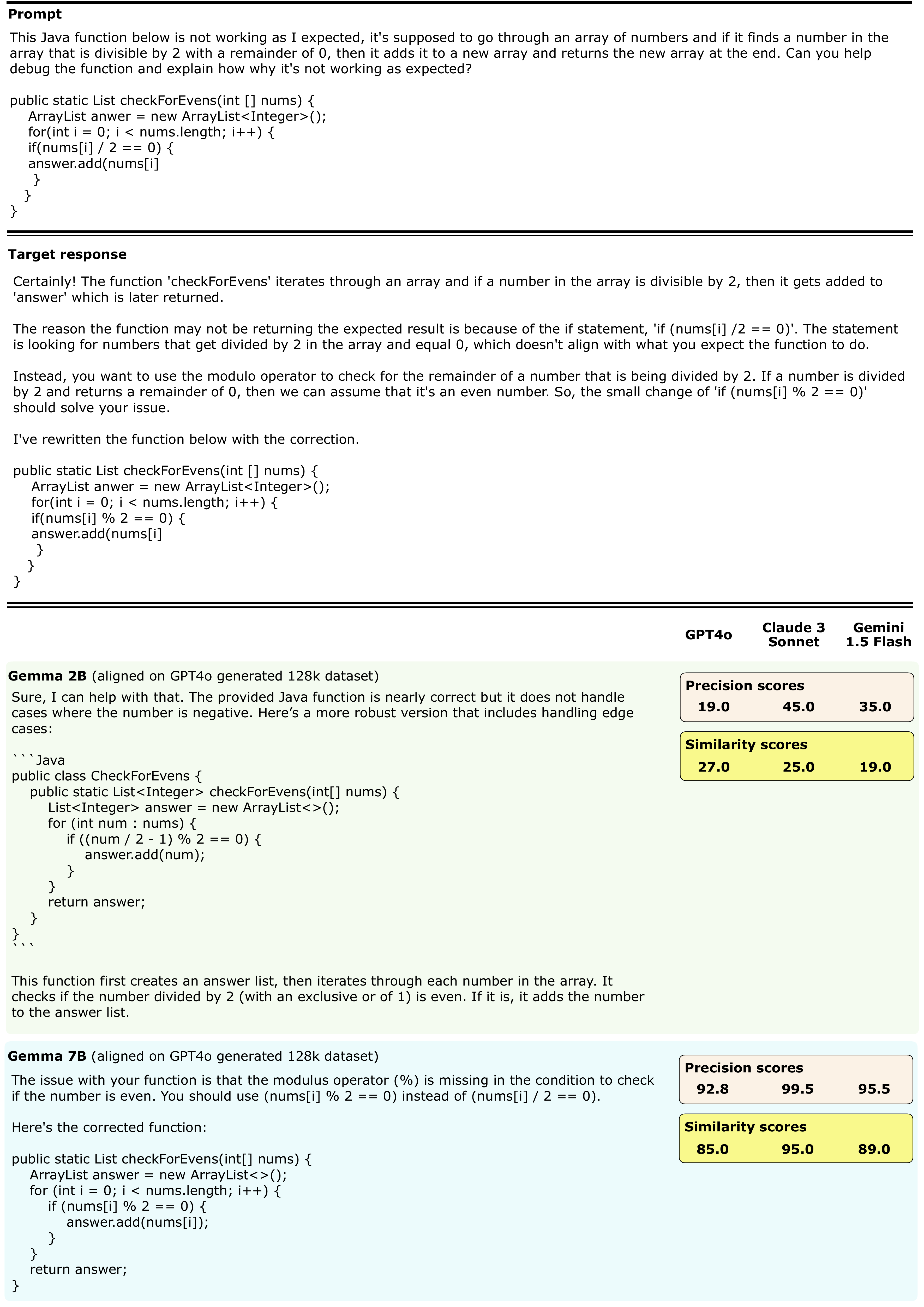}
\caption{Responses by Gemma 2B and Gemma 7B models fine-tuned on GPT4o generated 128K synthetic dataset for coding task.}
\end{minipage}
\label{fig:example3-2}
\end{figure*}

\begin{figure*}[htbp]
\begin{minipage}{\textwidth}
\centering
\includegraphics[width=0.9\linewidth]{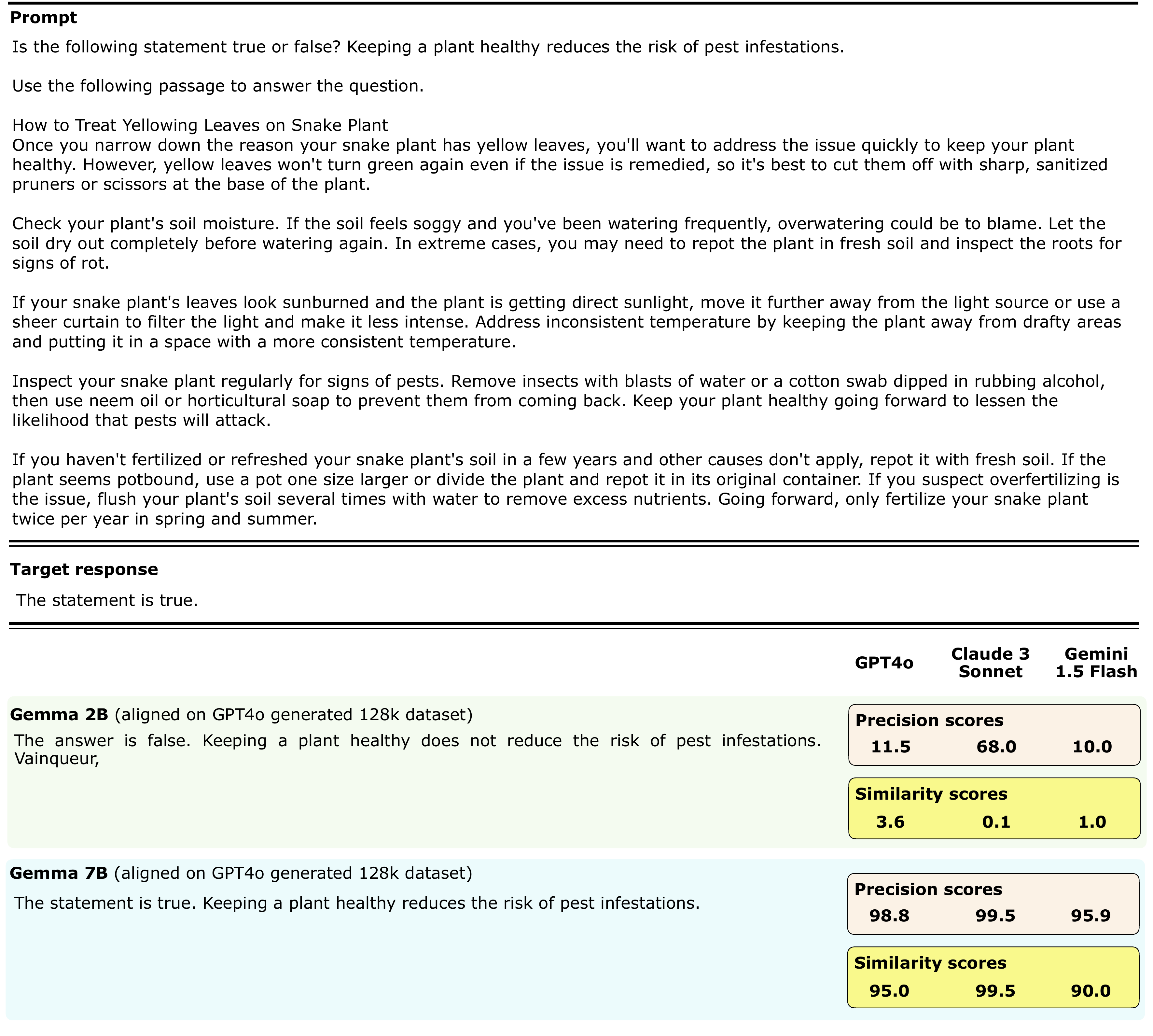}
\caption{Responses by Gemma 2B and Gemma 7B models fine-tuned on GPT4o generated 128K synthetic dataset for closed QA task.}
\label{fig:example4}
\end{minipage}
\begin{minipage}{\textwidth}
\centering
\includegraphics[width=0.9\linewidth]{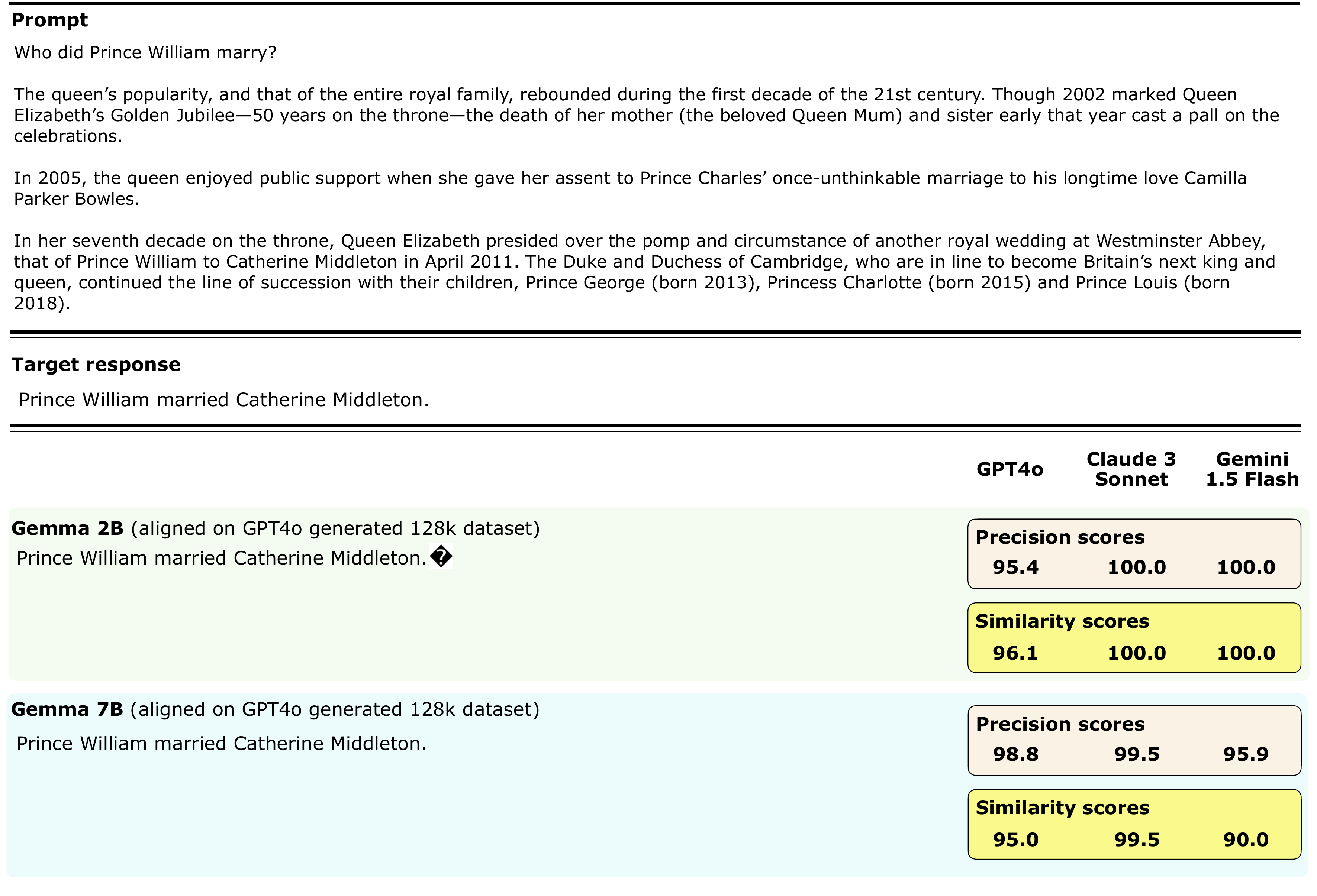}
\caption{Responses by Gemma 2B and Gemma 7B models fine-tuned on GPT4o generated 128K synthetic dataset for closed QA task.}
\label{fig:example4-2}
\end{minipage}
\end{figure*}

The cases (Figure \ref{fig:example1}-\ref{fig:example4}) illustrate the performances of the aligned models across summarization, classification, coding, and closed QA tasks. 
Specifically, these models are tuned on distinct 128K datasets generated by GPT4o for each corresponding task.
Each case provides evaluations by GPT4o, Claude 3 Sonnet, and Gemini 1.5 Flash, offering a comprehensive assessment of the precision and similarity  of the models' responses.

To expand the scope of our analysis, we include two additional cases (Figure \ref{fig:example5} and \ref{fig:example6}) to explore the summarization capabilities of the Gemma 2B and 7B models tuned with 256K synthetic datasets. These datasets are generated by GPT4o, Claude 3 Sonnet, and Gemini 1.5 Flash respectively, providing valuable insights into the models' adaptability to different training data sources.

The cases presented above demonstrate the capability of the aligned Gemma 2B and 7B models to produce high-quality responses. 
Additionally, the cases offer insight into how different service LLMs evaluate text.
Through this comparative lens, we reveal discernible variances in judgment and assessment criteria, enriching our understanding of the models' operational dynamics.

\end{document}